\documentclass[lettersize,journal]{IEEEtran}

\usepackage{amsmath,amssymb,amsfonts}
\DeclareMathOperator{\Enc}{Enc}
\DeclareMathOperator{\Dec}{Dec}

\usepackage{graphicx}
\usepackage[caption=false,font=normalsize,labelfont=sf,textfont=sf]{subfig}
\usepackage{stfloats}          %
\usepackage{placeins}          %

\usepackage{float}             %
\usepackage{cuted}
\usepackage{capt-of}

\usepackage{array,booktabs,multirow}
\usepackage[dvipsnames]{xcolor}
\usepackage{colortbl}
\usepackage{makecell}

\usepackage{algorithm}
\usepackage{algpseudocode}     %

\usepackage[utf8]{inputenc}
\usepackage{textcomp}
\usepackage{url}
\usepackage{pifont}
\usepackage[numbers]{natbib}
\usepackage{balance}           %

\definecolor{headercolor}{HTML}{2A9D8F}
\definecolor{rowalt}{HTML}{EEF7F6}
\definecolor{checkgreen}{HTML}{2A9D8F}
\definecolor{crossred}{HTML}{C0392B}
\newcommand{\cmark}{{\color{checkgreen}\ding{51}}}
\newcommand{\xmark}{{\color{crossred}\ding{55}}}

\usepackage{enumitem}
\setlist[enumerate]{leftmargin=*,labelsep=0.5em,itemsep=0pt,topsep=2pt}
\setlist[itemize]{leftmargin=*,labelsep=0.5em,itemsep=0pt,topsep=2pt}
\usepackage[hidelinks]{hyperref}

\begin{document}
\bstctlcite{IEEEexample:BSTcontrol}

\title{
Grounded in Contact: \\ Reducing Floating and Interpenetration Artifacts with Multimodal Skill-Effect Models
}

\author{Siyeon Kim$^{1}$, Mohanraj Devendran Shanthi$^{1,\dagger}$, Joseph C. Liechty$^{1,\dagger}$, and Tucker Hermans$^{1,2}$%
\thanks{$^{\dagger}$These authors contributed equally.}%
\thanks{$^{1}$University of Utah. $^{2}$NVIDIA Research.}%
\thanks{This work has been submitted to the IEEE for possible publication. Copyright may be transferred without notice, after which this version may no longer be accessible.}%
}

\maketitle

\begin{strip}
  \centering
  \includegraphics[width=\linewidth]{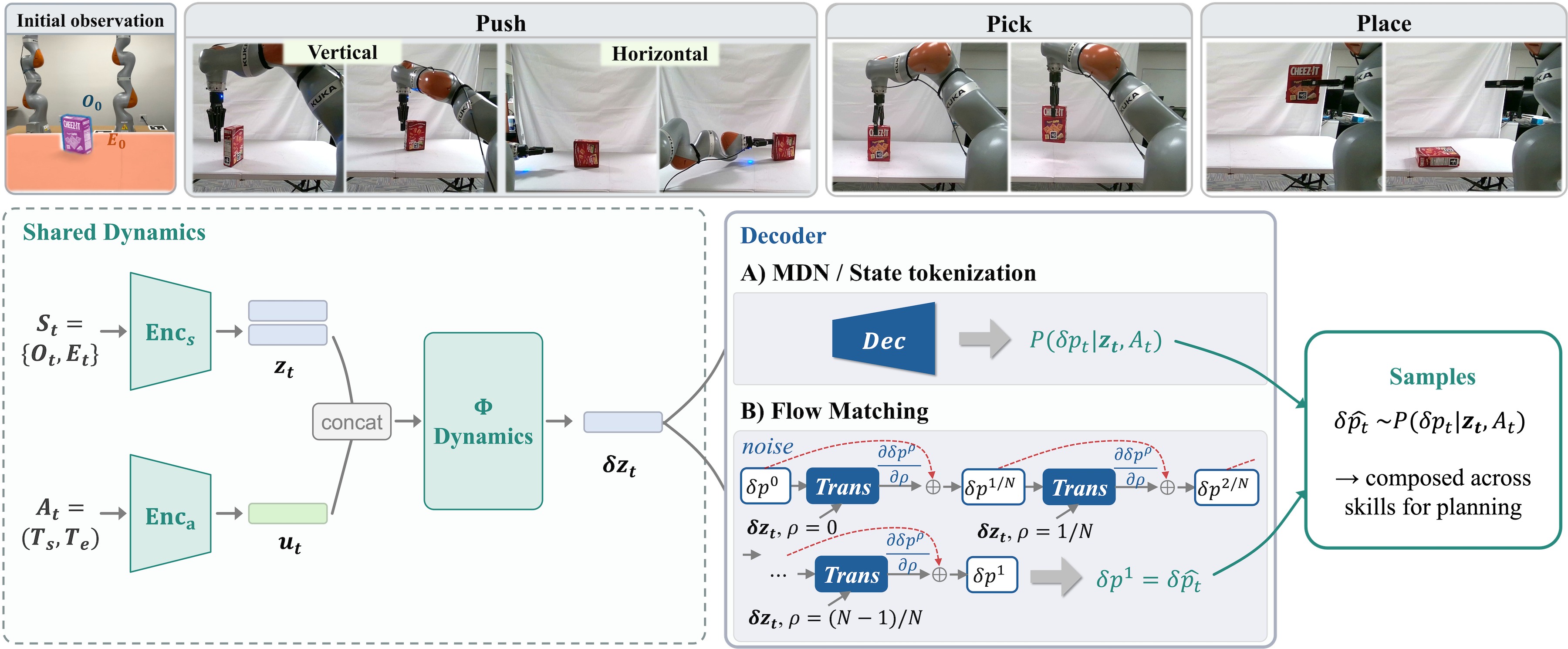}
  \captionof{figure}{\textbf{Multimodal probabilistic skill-effect models.}
  We encode segmented object point clouds and parameterized robot skills into latent state-action embeddings. A latent dynamics model $\Phi$ predicts single-step skill effects, which are decoded into distributions over $SE(3)$ object pose changes using one of three multimodal probabilistic decoders: mixture density networks, state tokenization, or flow matching. These models provide stochastic predictions that can be sampled or composed for planning under uncertainty across distinct skills.}
  \label{fig:comparison}
\end{strip}

\begin{abstract}
Learning to predict the effects of contact-rich manipulation skills requires models that represent distinct outcome modes rather than a single expected result. Even for basic skills such as pushing, picking, and placing, small variations in contact conditions or object geometry can produce qualitatively different object motions under nearly identical states and actions. Deterministic and Gaussian models tend to average over these alternatives or represent them as a single broad distribution, obscuring distinct modes such as sliding, toppling, and rotating. We study probabilistic skill-effect models that learn multimodal distributions over object pose changes given partial point clouds and actions. We instantiate this abstraction with three representative model classes: mixture density networks, state tokenization, and flow matching, each paired with latent dynamics for next-step prediction. Across diverse objects and skills, explicitly multimodal models predict which outcome occurs more accurately than deterministic and Gaussian baselines. In planning, flow matching significantly improves task success over the deterministic baseline on every skill and achieves the highest success on every multi-step sequence. On a real robot, probabilistic models succeed at pick about twice as often as the deterministic baseline, providing evidence that probabilistic skill effects offer a practical abstraction for reasoning and planning under uncertainty.
\end{abstract}

\begin{IEEEkeywords}
Uncertainty modeling, contact-rich manipulation, planning under uncertainty, latent world models, flow matching, robot manipulation.
\end{IEEEkeywords}

\section{Introduction}
\label{sec:introduction}

Learned manipulation models must generalize beyond their training data, yet contact-rich skills make this difficult: small changes in contact conditions, object geometry, or inertial properties can produce qualitatively different outcomes such as sliding, toppling, or rotating. Real-world deployment further introduces perception errors, motion execution errors~\citep{11-bayesian-grasp-planning}, and sim-to-real mismatch~\citep{22-sim2real-survey,19-bayessim,25-realissimbridgingsimtorealgap}, especially when models operate across objects from partial perceptual inputs such as images or point clouds. These factors do not merely broaden predictions around a single mean; they induce distinct, physically plausible outcome modes under nearly identical states and actions. We therefore argue for skill-effect models (SEMs) that explicitly represent multimodal robot action effects.

Robots with reliable action-outcome uncertainty estimates can act more safely~\citep{luo_multi-robot_2020} and efficiently~\citep{stephens_planning_2024}. For example, a planner that anticipates multiple possible effects can avoid pushes likely to topple a crate rather than slide it, preventing unsafe states and costly recovery actions. While probabilistic safety filters can mitigate some planning failures~\citep{luo_multi-robot_2020,stephens_planning_2024}, their effectiveness depends on the fidelity and structure of the underlying action-effect model. More generally, robots that reason about and visualize alternative futures can solicit human intervention under high uncertainty~\citep{ren2023knowno,chi2024uncertainty,alhaji_trust_2021} or communicate uncertainty to nearby humans, improving trust and transparency~\citep{alhaji_trust_2021,ullrich_development_2021,alhaji_trust_2025}.

Stochastic dynamics models have a long history in robot manipulation, including Gaussian process regression~\citep{deisenroth2015gaussian}, Gaussian mixture models of state-space dynamics~\citep{Khansari-Zadeh-tro2011-stable}, and neural network ensembles~\citep{chua-neurips2018-pets}. Because these methods model low-level transitions or represent uncertainty indirectly, they do not directly support reasoning about mode probabilities or visualizing alternative outcomes. SEMs~\citep{sharma2020relational} instead abstract away low-level dynamics by predicting the outcomes of temporally extended skills, and prior work shows that explicitly modeling skill-effect uncertainty with mixture density networks (MDNs)~\citep{bishop1994mixture} can improve planning under uncertainty~\citep{22-arxiv-uncertaintygoalplanning}. The MDN-based SEM in~\citep{22-arxiv-uncertaintygoalplanning}, however, assumes known object identities and trains separate models per object.

In this paper, we propose learning multimodal, probabilistic SEMs directly from partial point clouds. We train three representative classes of multimodal architectures: explicit mixture densities, state tokenization, and flow matching. We study their tradeoffs for predicting and reasoning about skill outcomes under uncertainty; see Figure~\ref{fig:comparison} and Table~\ref{tab:model_comparison}.
In summary, we make the following contributions: 
\begin{enumerate}
\item \textbf{Object-generalizable multimodal skill-effect models} --- We learn probabilistic skill effects from partial point clouds across seen and unseen objects, without object-specific models;
\item \textbf{Comparison of multimodal representations} --- We compare mixture densities, state tokenization, and flow matching against deterministic and Gaussian baselines on push, pick, and place; and
\item \textbf{Benefits of multimodality for prediction and planning} --- We show that the multimodal heads predict which outcome occurs more accurately than deterministic and single Gaussian baselines. In planning, flow matching significantly improves task success over the deterministic baseline on every skill in simulation, and on a real robot the probabilistic models succeed at pick about twice as often. Flow matching is the most accurate of these heads and the MDN the fastest to plan with.
\end{enumerate}

\begin{table}[t]
\centering
\caption{Properties of the output heads compared in this paper.}
\label{tab:model_comparison}
\setlength{\tabcolsep}{3pt}
\renewcommand{\arraystretch}{1.3}
\begin{tabular}{lcccc}
\arrayrulecolor{headercolor}
\rowcolor{headercolor}
\textcolor{white}{\textbf{}} & 
\textcolor{white}{\textbf{Det./Gauss.}} &
\textcolor{white}{\textbf{MDN}} & 
\textcolor{white}{\textbf{ST}} & 
\textcolor{white}{\textbf{FM}} \\
\rowcolor{rowalt}
Multimodality & \xmark & \cmark & \cmark & \cmark \\
Output space & Continuous & Continuous & Discrete & Continuous \\
\rowcolor{rowalt}
\raisebox{0.9ex}{Physical plausibility$^{*}$} & \shortstack{\rule{0pt}{2.6ex}\textbf{Medium}/\\{\color{crossred}\textbf{Lower}}} & \raisebox{0.9ex}{\color{checkgreen}\textbf{Higher}} & \raisebox{0.9ex}{\textbf{Medium}} & \raisebox{0.9ex}{\color{checkgreen}\textbf{Higher}} \\
Quantization-free & \cmark & \cmark & \xmark & \cmark \\
\rowcolor{rowalt}
Inference speed$^{\dagger}$ & \color{checkgreen}\textbf{Fast} & \color{checkgreen}\textbf{Fast} & \textbf{Moderate} & \color{crossred}\textbf{Slow} \\
\bottomrule
\end{tabular}\\[2pt]
\parbox{\columnwidth}{\footnotesize ST is state tokenization and FM is flow matching. $^{*}$Share of predicted object poses that are physically valid, resting on the table without floating, penetrating, or tilting, aggregated over skills (Figure~\ref{fig:prediction_quality}). $^{\dagger}$Time to predict outcomes for one planning batch of candidate actions (Appendix~\ref{sec:app_timing}).}
\end{table}

\section{Related Work}
\label{sec:related_works}
\textbf{Skill-Effect Models (SEMs).}
Learning SEMs enables robots to generalize to new tasks by composing basic skill primitives~\citep{kaelbling2017learning}. Prior work has leveraged SEMs to learn preconditions and effects of skill execution for task-and-motion planning~\citep{wang2021learningcompositionalmodelsrobot, liang2022searchbasedtaskplanninglearned, wang2018activemodellearningdiverse}, and to predict geometric effects such as pose changes during manipulation~\citep{paxton2021predicting, sharma2020relational, 24-arxiv-points2plans, huang2025fail2progresslearningrealworldrobot}. These geometric SEMs are typically deterministic, limiting their ability to represent the distinct outcome modes that arise in manipulation. MDN-based SEMs~\citep{22-arxiv-uncertaintygoalplanning} model outcome uncertainty, but assume known objects and train a separate model per object. We instead learn multimodal SEMs that generalize across objects from partial point clouds.

\textbf{Stochastic and Latent Dynamics.}
Stochastic dynamics models have long been used to represent transition uncertainty in robotics, including Gaussian processes~\citep{deisenroth2015gaussian}, Gaussian mixture models~\citep{Khansari-Zadeh-tro2011-stable}, and ensemble-based neural dynamics models~\citep{chua-neurips2018-pets}. Gaussian processes and ensembles improve uncertainty estimation but capture multimodality only indirectly, and these mixture models describe low-level state-space dynamics rather than temporally extended skill effects. Latent dynamics models, often called world models~\citep{ha2018world,hafner2020dream,hafner2021mastering,hafner2023dreamerv3}, learn stochastic transitions in a learned latent space for imagination-based control, but do not explicitly parameterize multimodal distributions over temporally extended skill effects.

\textbf{Discretization and Tokenization.}
Another approach to multimodal modeling discretizes continuous spaces and learns categorical distributions over tokens. This strategy has driven recent progress in robot policy learning through action tokenization, as in RT-1, RT-2, OpenVLA, and FAST~\citep{22-arxiv-rt1, 23-arxiv-rt2, 24-arxiv-openvla,25-arxiv-fastefficientactiontokenization}. Related state-space tokenization methods have also emerged in trajectory forecasting~\citep{23-iccv-motionlm, 24-iclr-trajeglish, 24-arxiv-closedlooptokenized}. Inspired by these works, we adopt state tokenization for skill effects.

\textbf{Flow-Based and Generative Models.}
Flow-matching and diffusion models offer powerful mechanisms for representing multimodal distributions~\citep{lipman_flow_2023, ho_denoising_2020}. Often paired with transformer backbones~\citep{peebles_scalable_2023}, these models have achieved strong performance in robot action policies~\citep{chi_diffusion_2025, black_0_2024} and world modeling~\citep{alonso_diffusion_2024, hafner_training_2025}. Flow matching in particular learns deterministic vector fields that transport probability mass, enabling efficient sampling~\citep{liu_flow_2022}. At the skill level, Generative Skill Chaining~\citep{mishra2023generative} learns a diffusion model of each skill's transitions and composes these models to generate long-horizon plans. However, generative models have not been systematically compared with mixture-density and tokenized alternatives for predicting skill-effect distributions from partial object observations.

\begin{figure*}[t]
\centering
\includegraphics[width=\textwidth]{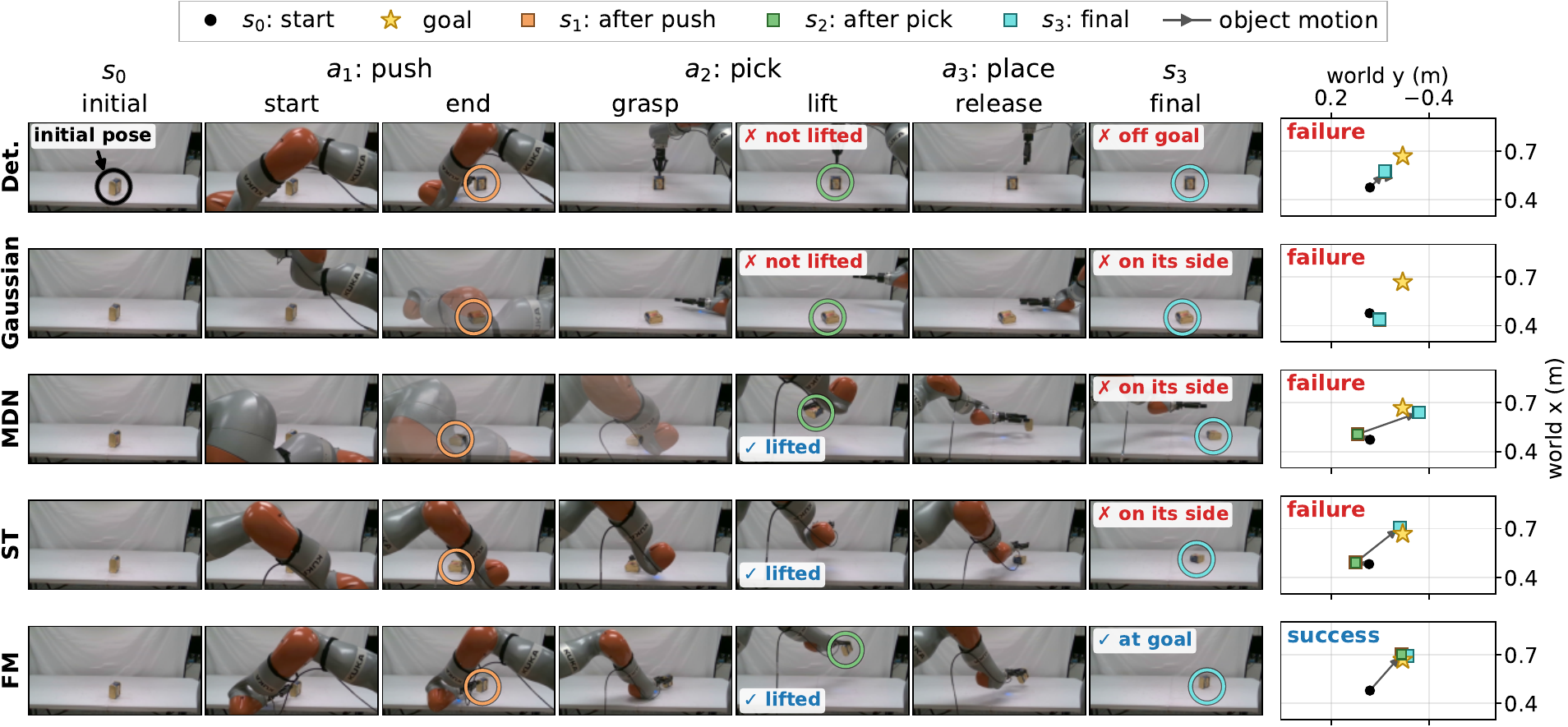}
\caption{Push--pick--place on the real robot with the seen potted meat can, from the same start and goal for every model, one run each. Models are the deterministic model (Det.), the Gaussian, the MDN, state tokenization (ST), and flow matching (FM). Each model plans the whole sequence with its own predicted outcomes, so the models choose different actions and reach different final poses. Left: camera views at each step. In frames where the robot occludes the object, it is rendered semi-transparent by blending with a frame without the robot. Right: the object's motion from the start (circle) to the final pose. A run succeeds under the criterion of Eq.~\eqref{eq:success} at $\epsilon_{\mathrm{pos}}=5$\,cm and $\epsilon_{\mathrm{rot}}=20^\circ$, and failure labels indicate which criterion was not met.}
\label{fig:real_multistep}
\end{figure*}

\section{Multimodal Probabilistic Skill-Effect Models}
\label{sec:methods}
We seek to learn a probabilistic SEM that predicts the distribution of object motion induced by a parameterized manipulation skill. At time \(t\), the model observes a pair of partial-view point clouds,
\(
S_t = \{O_t, E_t\},
\)
where \(O_t\) denotes the target-object point cloud and \(E_t\) denotes the surrounding environment. Given this observation and an action \(A_t\), the model predicts a distribution over the relative object pose change, following~\citep{24-arxiv-points2plans},
\(
\delta p_t = p_t^{-1}\,p_{t+1} \in SE(3),
\)
where \(p_t\) and \(p_{t+1}\) are the world-frame object poses before and after executing the action. We denote the learned predictive distribution as
\(
P(\delta p_t \mid \mathbf{z}_t, A_t),
\)
where \(\mathbf{z}_t\) is the latent representation of \(S_t\) (Section~\ref{sec:network}).
This distribution induces a prediction over the next world-frame object pose through
\(
p_{t+1}
=
p_t\,\delta p_t.
\)

Each action is written as
\(
A_t = (a_t,\boldsymbol{\theta}_t),
\)
where \(a_t\) is a discrete manipulation skill, such as push, pick, or place, and \(\boldsymbol{\theta}_t\) contains the continuous skill parameters. This formulation treats skills as temporally extended actions with stochastic effects, following prior work on SEMs~\citep{liang2022searchbasedtaskplanninglearned,24-arxiv-points2plans}. Rather than predicting a single deterministic outcome, our goal is to learn the full conditional distribution over possible effects, allowing the model to represent distinct contact-induced modes such as sliding, rotating, toppling, or no motion.

Given a learned SEM, the robot can plan by selecting an \(H\)-step sequence of actions, one skill per step,
\(
\tau = (A_0,\ldots,A_{H-1})
\)
that minimizes expected cost under the learned predictive dynamics:
\begin{equation}
\tau^\star
=
\arg\min_{\tau}
\mathbb{E}_{\delta p_{0:H-1} \sim P}
\left[
c(p_H,P_G)
\right]
\end{equation}
subject to
\begin{equation}
p_{t+1} = p_t\,\delta p_t,
\qquad
\delta p_t \sim P(\delta p_t \mid \mathbf{z}_t,A_t).
\end{equation}
Here \(P_G\) denotes the goal distribution, and \(c(\cdot)\) is a task-dependent cost.

In the remainder of this section, we describe the model architecture used to represent \(P(\delta p_t \mid \mathbf{z}_t,A_t)\), the training objectives for each probabilistic decoder, and the planning procedure that uses the learned SEM.

\subsection{Model Architecture}
\label{sec:network}

We model the probabilistic SEM
\(
P(\delta p_t \mid \mathbf{z}_t,A_t)
\)
using three components: encoders \(\Enc\), a latent dynamics model \(\Phi\), and a decoder \(\Dec\) (the output head). The encoders map partial point-cloud observations and parameterized actions into a shared latent state-action representation. The latent dynamics model predicts the skill-induced change in this latent space, and the decoder maps the predicted latent change to a distribution over relative object pose changes.

\paragraph{Encoders}
We encode both the scene observation and the action \(A_t\). Given the input point cloud, we first segment the target object \(O_t\) from the surrounding environment \(E_t\) and extract segment-specific feature vectors. We use Point Transformer v3~\citep{Wuetal2024} as the point-cloud encoder,
\(
\mathbf{x}_t^O = \Enc_{\mathrm{pc}}(O_t),
\quad
\mathbf{x}_t^E = \Enc_{\mathrm{pc}}(E_t).
\)
To distinguish the two segments, we concatenate each feature with a learned segment embedding, yielding
\(
\mathbf{z}_t^O = \mathbf{x}_t^O \oplus \mathbf{e}^O
\)
and
\(
\mathbf{z}_t^E = \mathbf{x}_t^E \oplus \mathbf{e}^E.
\)
The latent scene representation is
\(
\mathbf{z}_t = \{\mathbf{z}_t^O, \mathbf{z}_t^E\},
\)
and we denote this scene encoder by \(\mathbf{z}_t = \Enc_s(S_t)\).

We assume a set of parameterized skills, each of which takes continuous parameters \(\boldsymbol{\theta}_t\) and executes a robot trajectory. In this paper, we use horizontal push, vertical push, pick, and place (Figure~\ref{fig:comparison}).
Each skill is parameterized by \(\boldsymbol{\theta}_t\in\mathbb{R}^{18}\), the end-effector poses at the start and end of the skill, each represented by a 9D vector: a 3D translation and a continuous 6D rotation representation formed by the first two columns of the rotation matrix~\citep{19-cvpr-ortho6d}.
The action encoder maps these parameters to an action embedding,
\(
\mathbf{u}_t = \Enc_a(A_t).
\)
We train one model per skill, using the same architecture for all skills.

\paragraph{Latent dynamics model}
Given the latent scene representation \(\mathbf{z}_t\) and action embedding \(\mathbf{u}_t\), our transformer-based latent dynamics model~\citep{23-transformer} predicts the relative change in latent space:
\(
\delta\mathbf{z}_t = \Phi(\mathbf{z}_t,\mathbf{u}_t).
\)
The predicted next latent state is
\(
\hat{\mathbf{z}}_{t+1} = \mathbf{z}_t + \delta\mathbf{z}_t,
\)
similar to a prior latent SEM~\citep{24-arxiv-points2plans}. However, unlike previous work, the latent update \(\delta\mathbf{z}_t\) is not decoded into a single pose; the decoder maps it to a distribution over pose changes (MDN, state tokenization) or samples from that distribution (flow matching).

\paragraph{Decoder}
The decoder maps the predicted latent change \(\delta\mathbf{z}_t\) to a relative object pose change,
\(
\delta p_t \in SE(3).
\)
For the deterministic baseline~\citep{24-arxiv-points2plans}, the decoder outputs a single pose estimate,
\(
\widehat{\delta p}_t
=
\Dec_{\mathrm{det}}(\delta\mathbf{z}_t).
\)
This transform is applied to the current object pose or point cloud to predict the post-action object state.
To model multimodal outcomes of contact-rich interaction, such as sliding, rotating, toppling, or no motion, we replace the deterministic decoder with one of three probabilistic decoders with parameters \(\psi\):
\(
P(\delta p_t \mid \mathbf{z}_t,A_t)
=
\Dec_\psi(\delta\mathbf{z}_t).
\)

\begin{enumerate}
    \item \textit{\textbf{Mixture Density Networks (MDN)}} represent the distribution as a mixture of \(K\) Gaussians over relative pose changes:
    \begin{multline*}
    P(\delta p_t \mid \mathbf{z}_t,A_t)
    = \sum_{k=1}^{K} \alpha_k(\mathbf{z}_t,A_t;\psi)\\
    \cdot\,\mathcal{N}\bigl(\delta p_t \mid \boldsymbol{\mu}_k(\mathbf{z}_t,A_t;\psi),\,\Sigma_k(\mathbf{z}_t,A_t;\psi)\bigr).
    \end{multline*}
    The decoder predicts the mixture weights \(\alpha_k\), component means \(\boldsymbol{\mu}_k\), and covariances \(\Sigma_k\) from \(\delta\mathbf{z}_t\). A softmax constrains the mixture weights to sum to one, and the covariance parameterization is chosen to ensure positive definiteness~\citep{bishop1994mixture,22-arxiv-uncertaintygoalplanning}. Each component is a Gaussian on \(\mathbb{R}^3\times SO(3)\): Euclidean in translation and defined in the tangent space of \(SO(3)\) in rotation~\citep{barfoot2014associating}, with the mean rotation predicted in the continuous 6D representation~\citep{19-cvpr-ortho6d}. With \(K=1\), the mixture reduces to a single Gaussian, which we use as the Gaussian baseline. 

    \item \textit{\textbf{State Tokenization (ST)}} discretizes the continuous relative-pose space into translation and rotation token vocabularies, similar to~\citep{24-iclr-trajeglish,23-iccv-motionlm}. We use the \textit{K-disks} algorithm~\citep{24-iclr-trajeglish} and extend it to \(SE(3)\) by constructing separate vocabularies for translation and rotation,
    \(
    \mathcal{V}^{\mathrm{pos}}
    \)
    and
    \(
    \mathcal{V}^{\mathrm{rot}}
    \) (Figure~\ref{fig:token_sets}, Appendix~\ref{sec:app_hyper}).
    The decoder predicts categorical distributions over tokens \(v^{\mathrm{pos}}\in\mathcal{V}^{\mathrm{pos}}\) and \(v^{\mathrm{rot}}\in\mathcal{V}^{\mathrm{rot}}\),
    \(
    P(v^{\mathrm{pos}},v^{\mathrm{rot}}\mid \mathbf{z}_t,A_t)
    =
    \Dec_{\mathrm{tok}}(\delta\mathbf{z}_t),
    \)
    reducing pose prediction to two multiclass classification problems. We use vocabularies of
    \(
    |\mathcal{V}^{\mathrm{pos}}| = |\mathcal{V}^{\mathrm{rot}}| = 512
    \)
    tokens, selected by comparing vocabularies of 128 to 512 tokens on the validation set (Appendix~\ref{sec:app_hyper}), choosing clustering thresholds to balance coverage of the observed transition space against sparsity in rarely used tokens. Validation and test transitions are assigned to their nearest translation and rotation tokens.

    \item \textit{\textbf{Flow Matching (FM)}}~\citep{lipman_flow_2024} parameterizes a continuous generative model over relative pose changes on the \(SE(3)\) manifold~\citep{chen_flow_2024}. The flow decoder learns a time-dependent vector field conditioned on the predicted latent dynamics:
    \(
    \frac{\partial\,\delta p^\rho}{\partial\rho}
    =
    \Dec_{\mathrm{flow}}
    \left(
    \delta p^\rho,
    \rho,
    \delta\mathbf{z}_t
    \right),
    \)
    where \(\rho\in[0,1]\) denotes the flow time and \(\delta p^\rho\) is the current partially denoised pose sample. The network predicts a twist, which is integrated on \(SE(3)\) to transport samples from a base distribution to the learned conditional distribution over skill effects. The architecture uses a transformer backbone with adaptive layer normalization to condition on the flow time, similar to Diffusion Transformers~\citep{peebles_scalable_2023}. At inference time, we sample an initial pose from the base distribution and integrate the learned vector field with the midpoint method for five steps.
\end{enumerate}

\subsection{Training}
\label{sec:training}

We train the encoder \(\Enc\), latent dynamics model \(\Phi\), and decoder \(\Dec\) end-to-end with a decoder-specific task loss, which supervises either the predicted relative pose distribution or, for flow matching, the learned velocity vector field. The deterministic baseline is trained as in Points2Plans~\citep{24-arxiv-points2plans}, with a geodesic pose loss and their latent transition loss (Appendix~\ref{sec:app_training}). The probabilistic models are trained with their task loss alone, without the latent transition loss, which degrades their predictions (Appendix~\ref{sec:app_reg}).

\paragraph{Relative pose target}
For all pose-prediction decoders, the target is the relative object pose change \(\delta p_t=(\delta \mathbf{t}_t,\delta R_t)\), with translation \(\delta \mathbf{t}_t\) and rotation \(\delta R_t\). Since the object pose is not observable from a partial point cloud, we define the object frame \(p_t\) from the observation: object origin is the bottom center of the object (the point on the support surface below the center of the object cloud), object \(z\)-axis is the world up axis, and object yaw is the principal axis of the cloud's horizontal projection, oriented away from the robot base. \(p_{t+1}\) is this frame carried rigidly with the object, so \(\delta p_t\) is the object motion expressed in the frame observed at time \(t\).

\paragraph{Mixture-density decoder}
We train the MDN decoder by minimizing the negative log-likelihood of the observed transition,
\begin{equation*}
\mathcal{L}_{\mathrm{MDN}}
= -\log \sum_{k=1}^{K} \alpha_k\,\mathcal{N}(\delta p_t \mid \boldsymbol{\mu}_k,\Sigma_k).
\end{equation*}
Each component factorizes into translation and rotation terms,
\begin{equation*}
\begin{aligned}
\mathcal{N}(\delta p_t \mid \boldsymbol{\mu}_k,\Sigma_k)
={}& \mathcal{N}\bigl(\delta \mathbf{t}_t \mid \boldsymbol{\mu}^{\mathrm{pos}}_k,\Sigma^{\mathrm{pos}}_k\bigr)\\
&\cdot\,\mathcal{N}\bigl(\mathrm{Log}(\bar R_k^{\top}\delta R_t) \mid \mathbf{0},\Sigma^{\mathrm{rot}}_k\bigr),
\end{aligned}
\end{equation*}
where \(\boldsymbol{\mu}_k=(\boldsymbol{\mu}^{\mathrm{pos}}_k,\bar R_k)\), \(\Sigma_k=\mathrm{diag}(\Sigma^{\mathrm{pos}}_k,\Sigma^{\mathrm{rot}}_k)\), and \(\mathrm{Log}\) and its inverse \(\mathrm{Exp}\) are the logarithm and exponential maps of \(SO(3)\). The first term is a Euclidean Gaussian in translation, and the second is a Gaussian on a tangent vector in \(\mathbb{R}^3\). This factorization assumes no within-component covariance between translation and rotation, while the shared mixture weights still allow each component to represent a coupled pose-change mode.

\paragraph{State-tokenization decoder}
For state tokenization, we assign the ground-truth relative pose change to translation and rotation tokens,
\(
v_t^{\mathrm{pos}} = q^{\mathrm{pos}}(\delta \mathbf{t}_t),
\qquad
v_t^{\mathrm{rot}} = q^{\mathrm{rot}}(\delta R_t),
\)
where \(q^{\mathrm{pos}}\) and \(q^{\mathrm{rot}}\) map continuous targets to their nearest vocabulary elements. The token decoder predicts independent categorical distributions over the two vocabularies,
\(
P(v^{\mathrm{pos}},v^{\mathrm{rot}}\mid \mathbf{z}_t,A_t)
=
P(v^{\mathrm{pos}}\mid \mathbf{z}_t,A_t)\,P(v^{\mathrm{rot}}\mid \mathbf{z}_t,A_t),
\)
both predicted by \(\Dec_{\mathrm{tok}}(\delta\mathbf{z}_t)\).
We train the model with the sum of translation-token and rotation-token cross-entropies:
\(
\mathcal{L}_{\mathrm{tok}}
=
-
\log P(v_t^{\mathrm{pos}}\mid \mathbf{z}_t,A_t)
-
\log P(v_t^{\mathrm{rot}}\mid \mathbf{z}_t,A_t).
\)

\paragraph{Flow-matching decoder}
The flow-matching decoder learns a conditional vector field on \(SE(3)\). For each training transition, we draw a base-distribution sample \(\delta p^0=(\delta\mathbf{t}^0,\delta R^0)\sim\mathcal{N}(\mathbf{0},\mathbf{I})\) in SE(3) tangent space, sample a flow time \(\rho\in[0,1]\), and move toward the ground-truth relative pose \(\delta p_t\) along a geodesic on \(\mathbb{R}^3\times SO(3)\): a straight line in translation and a geodesic in rotation,
\(
\delta\mathbf{t}^\rho = \delta\mathbf{t}^0 + \rho\,(\delta\mathbf{t}_t-\delta\mathbf{t}^0),
\qquad
\delta R^\rho = \delta R^0\,\mathrm{Exp}(\rho\,\boldsymbol{\omega}_t),
\qquad
\boldsymbol{\omega}_t=\mathrm{Log}\big((\delta R^0)^{\top}\delta R_t\big),
\)
so the target velocity \(\xi_t=[\delta\mathbf{t}_t-\delta\mathbf{t}^0,\ \boldsymbol{\omega}_t]^\top\) is constant in \(\rho\) \citep{chen_flow_2024,bose_se3-stochastic_2024,yim_fast_2023}. The decoder predicts an on-manifold twist,
\(
\widehat{\xi}_t^\rho
=
\Dec_{\mathrm{flow}}
\left(
\delta p^\rho,
\rho,
\delta \mathbf{z}_t
\right),
\qquad
\widehat{\xi}_t^\rho=[\widehat{\mathbf{v}}_t^\rho,\widehat{\boldsymbol{\omega}}_t^\rho]^\top ,
\)
which is trained to match the target flow velocity \(\xi_t\):
\(
\mathcal{L}_{\mathrm{flow}}
=
\left\|
\xi_t-\widehat{\xi}_t^\rho
\right\|_2^2 .
\)
We did not use optimal-transport pairing~\citep{tong_improving_2024} because doing so per condition for continuous conditioning variables collapses to no pairing. Additional implementation details, including optimization hyperparameters, loss weights, token construction, and flow integration settings, are provided in Appendix~\ref{sec:app_training}.

\subsection{Planning}
\label{sec:planning}

We use the learned SEM for model-predictive planning over parameterized skills. Given an initial observation \(S_0\), an initial object pose \(p_0\), and a goal distribution \(P_G\)~\citep{22-arxiv-uncertaintygoalplanning}, the planner searches for an action sequence
\(
\tau=(A_0,\ldots,A_{H-1})
\)
that minimizes the expected cost of the final object pose under the learned predictive distribution:
\begin{align}
\tau^\star
&=
\arg\min_{\tau\in\mathcal A^H}
J(\tau),
\\
J(\tau)
&=
\mathbb{E}_{\delta p_{0:H-1}\sim P}
\left[
c(p_H,P_G)
\right],
\label{eq:planning_objective}
\end{align}
subject to the rollout dynamics
\begin{align}
\delta\mathbf{z}_t &= \Phi\big(\Enc_s(S_t),\Enc_a(A_t)\big), \label{eq:planning_latent_delta}\\
\delta p_t &\sim \Dec_\psi(\delta\mathbf{z}_t), \label{eq:planning_pose_sample}\\
p_{t+1} &= p_t\,\delta p_t, \label{eq:planning_pose_rollout}\\
A_t &=(a_t,\boldsymbol{\theta}_t)\in\mathcal A, \label{eq:planning_action_space}\\
\boldsymbol{\theta}_{\min} &\leq \boldsymbol{\theta}_t \leq \boldsymbol{\theta}_{\max},
\qquad t=0,\ldots,H-1 .
\label{eq:planning_action_bounds}
\end{align}
Here \(\mathcal{A}\) is the action space, \(\boldsymbol{\theta}_{\min}\) and \(\boldsymbol{\theta}_{\max}\) bound the skill parameters, \(\delta\mathbf{z}_t\) is the predicted latent update for the target object, and \(\Dec_\psi\) denotes either a deterministic decoder or one of the probabilistic decoders described above. The cost \(c(p_H,P_G)\) measures task progress toward the goal. When the goal is represented as a distribution \(P_G\), we use the negative log likelihood
\[
c(p_H,P_G)=-\log P_G(p_H).
\]

For a candidate action sequence \(\tau\), the expectation in Eq.~\eqref{eq:planning_objective} is approximated by Monte Carlo rollouts. We draw \(M\) independent outcome sequences from the learned SEM,
\[
\delta p_t^{(m)}
\sim
P(\delta p_t\mid \mathbf{z}_t^{(m)},A_t;\psi),
\qquad
m=1,\ldots,M,
\]
roll out the corresponding object poses using Eq.~\eqref{eq:planning_pose_rollout}, and estimate the trajectory cost as
\begin{equation}
\widehat J(\tau)
=
\frac{1}{M}
\sum_{m=1}^{M}
c\big(p_H^{(m)},P_G\big).
\label{eq:mc_planning_cost}
\end{equation}
For the deterministic baseline, this estimate reduces to a single rollout with \(M=1\).

We optimize Eq.~\eqref{eq:planning_objective} using the cross-entropy method (CEM)~\citep{12-cem-planner, chua-neurips2018-pets}, summarized in Algorithm~\ref{alg:cem} (Appendix~\ref{sec:app_planner}). CEM maintains a distribution over continuous skill parameters, samples candidate action sequences, evaluates each candidate using Eq.~\eqref{eq:mc_planning_cost}, and updates the sampling distribution using the lowest-cost elite candidates. We initialize the sampling distribution as a uniform distribution over the skill-parameter ranges used for data collection, and discard candidates that leave the workspace, collide with the observed point cloud, or are kinematically infeasible. In single-step experiments, we use $H=1$, and in multi-step experiments, $H$ equals the number of skills in the skeleton.

\section{Experimental Results and Discussion}
\label{sec:result}

We compare our three multimodal SEMs, an MDN with five components, state tokenization (ST), and flow matching (FM), against two baselines, a deterministic regressor and a Gaussian. We also train each model with three seeds (Appendix~\ref{sec:app_qual}). As the deterministic baseline, we adopt the latent dynamics model of Points2Plans~\citep{24-arxiv-points2plans}, which directly regresses the relative object pose change in $SE(3)$. We evaluate three things: the quality of pose predictions in simulation, planning in simulation, and planning in the real world. The appendices are provided as supplementary material.

\subsection{Datasets}
\label{sec:datasets}
We collected push, pick, and place data in Isaac Sim~\citep{NVIDIA_Isaac_Sim} with a bimanual setup of two KUKA LBR iiwa arms, each with a Robotiq 2F-140 gripper. The training set covers 94 objects, 44 from YCB~\citep{15-ycb-dataset} and 50 primitive shapes, and each skill uses a subset of them. For each object, we executed nominal skills from stable resting poses. We then perturb with noise either the object pose or the skill parameters and re-execute in order to obtain failed executions: about 17\% of the push examples barely move the object, and about 40\% of the pick examples fail to lift it. We held out 20\% of the training-object episodes for validation, leaving 48{,}899 to 103{,}804 training episodes per skill. The test set contains new episodes of up to 11 training objects and up to 12 unseen objects from Google Scanned Objects (GSO)~\citep{gso}. Appendix~\ref{sec:app_training} lists the objects, perturbation ranges, and episode counts.

\subsection{Evaluation Metrics}
\label{sec:eval_metrics}

\textbf{Outcome-class Brier score} measures whether a model predicts which outcome occurs. Evaluating at the level of discrete outcomes follows work on trajectory prediction, where distance-based metrics do not necessarily capture performance in predicting interaction modes~\citep{hugenholtz2025modecollapse}. We assign each predicted sample and the observed outcome to one of three classes: keep, topple, and off for push and place, and held, ungrasped, and off for pick, where off contains physically invalid outcomes (Appendix~\ref{sec:app_eval}). The score is the squared error between the fraction of samples in each class and the observed class, summed over classes and averaged over episodes. It ranges from 0 to 2, and lower is better. 

\textbf{Physical plausibility} is the share of predicted outcomes in which the object is neither floating above the table, penetrating it, nor tilted away from its rest poses (Appendix~\ref{sec:app_eval}).

\textbf{Energy score (ES)}~\citep{gneiting2007strictly} is a sample-based scoring rule that measures how well the predicted distribution matches the observed outcome, with a lower score indicating a closer match. We compute it separately for position and orientation (Appendix~\ref{sec:app_eval}). 

\textbf{Calibration error} measures whether the predicted spread matches how much the outcomes actually vary: under a calibrated model, the observed outcome is indistinguishable from one more predicted sample. It is 0 for a perfectly calibrated model and, like the ES, is computed separately for position and orientation (Appendix~\ref{sec:app_eval}).

\textbf{Task success} measures whether the final object pose is within a position tolerance \(\epsilon_{\mathrm{pos}}\) and a rotation tolerance \(\epsilon_{\mathrm{rot}}\) of the goal (Appendix~\ref{sec:app_eval}).

\textbf{A significant difference} between two models is one whose 95\% object-level bootstrap interval of the paired difference excludes zero.

\begin{figure}[t]
  \centering
  \includegraphics[width=\columnwidth]{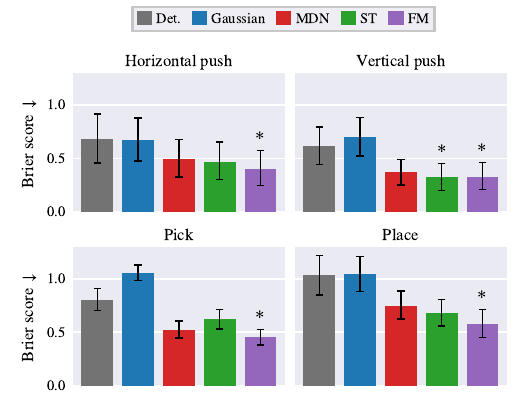}
  \caption{Outcome-class Brier score on the test set (lower is better). Error bars show 95\% object-level bootstrap intervals, and an asterisk marks the best model in each panel and the models not significantly different from it. ST and FM denote state tokenization and flow matching.}
  \label{fig:brier}
\end{figure}

\begin{figure*}[t]
  \centering
  \includegraphics[width=0.95\textwidth]{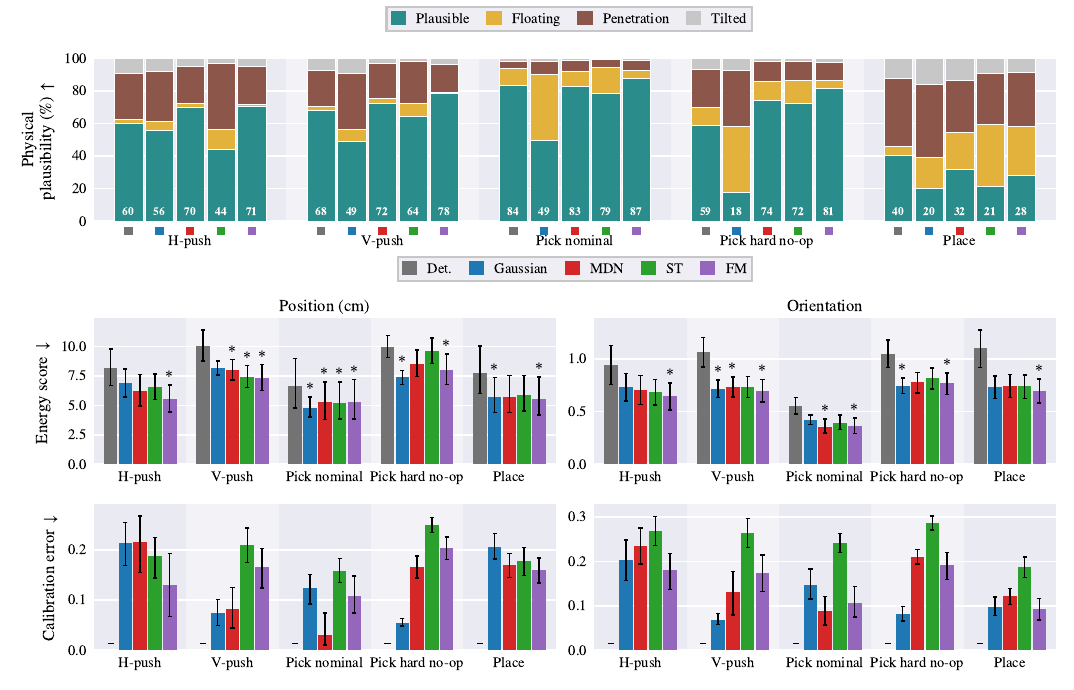}
  \caption{Prediction quality on the test set, with \(M{=}100\) samples per episode and one for the deterministic model. Pick hard no-op is a pick in which the gripper contacts the object but the pick fails.
  \textbf{Top.} Share of samples that are plausible, floating, penetrating, or tilted, with colored squares marking the models. The ground truth is at least 97.9\% plausible in every setting.
  \textbf{Middle and bottom.} Energy score and calibration error for position (left) and orientation (right), where lower is better. Calibration error is undefined for the deterministic model, which outputs a single sample.
  Error bars show 95\% CIs from an object-level bootstrap. In the energy-score panels, an asterisk marks the best model and the models not significantly different from it.}
  \label{fig:prediction_quality}
\end{figure*}

\subsection{Simulation Pose Predictions}
\begin{figure*}[t]
  \centering
  \includegraphics[width=\textwidth]{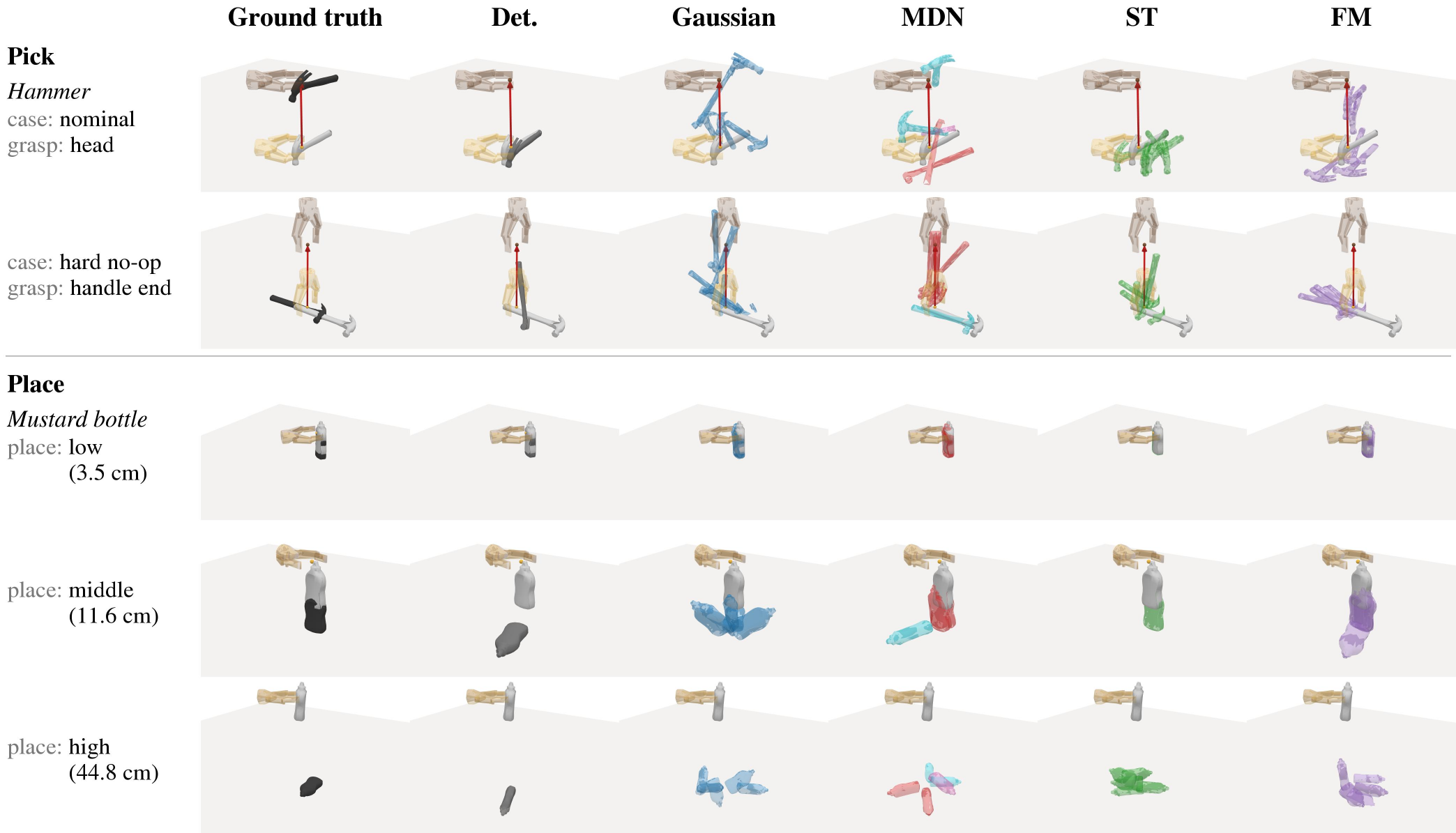}
  \caption{Predicted outcomes on pick and place. Each row is one episode, and each column shows the ground truth or the prediction of each model (ten samples per stochastic model, MDN colored by mixture component). The start pose is light gray, the ground-truth outcome is black, and the gripper start and end poses are yellow and brown, respectively, where applicable. Additional episodes are in Appendix~\ref{sec:app_qual}.}
  \label{fig:qual_pp}
\end{figure*}

\textbf{Multimodal decoders better predict which outcome occurs compared to deterministic and Gaussian baselines.} On all four skills, all three multimodal decoders achieved a significantly lower (better) Brier score than the deterministic and Gaussian models (Figure~\ref{fig:brier}), and flow matching scored the lowest for all skills except vertical push, where it was close to state tokenization. The Gaussian did not consistently improve on the deterministic model and was significantly worse on pick. 
These findings held across training seeds. In each of three seeds and on every skill, all three multimodal decoders had a lower Brier score than both baselines (Appendix~\ref{sec:app_qual}).

The multimodal models also generally produced more physically plausible predictions than the Gaussian (Figure~\ref{fig:prediction_quality}, top). The gap was largest on pick, where many of the Gaussian samples floated between the table and the gripper, while the multimodal models mostly kept the object on the table or in the gripper. On place, the deterministic model was the most plausible (40.3\%) because it rarely floated (5.5\%), whereas the probabilistic models floated in 20--38\% of their samples. 

The gap on pick arises where a small change in the grasp switches the outcome (Fig.~\ref{fig:qual_pp}, with more examples in Appendix~\ref{sec:app_qual}). The hammer grasped near its head is lifted, yet the deterministic model predicts that it stays on the table, and the Gaussian places its samples in between, while the MDN and flow matching place samples on both outcomes. Grasping the end of the handle instead leaves the hammer on the table. The single Gaussian again places samples between the two outcomes, which explains the Brier gap on pick. Place shows the same effect: the mustard bottle released from 11.6\,cm lands upright, but the deterministic model commits to a toppled pose, while the multimodal models represent both the upright and the toppled outcomes.

The energy score and calibration error measure a different property, how well the whole predicted distribution matches the outcome (Figure~\ref{fig:prediction_quality}, middle and bottom). All probabilistic models attained lower energy scores than the deterministic model, but the multimodal models did not consistently improve on the Gaussian. This is expected, because both metrics are only weakly sensitive to the number of outcome modes. The energy score barely distinguishes a two-component mixture from a single Gaussian with the same mean and covariance~\citep{marcotte2023regions}, and our calibration measure checks the spread of the samples rather than which outcomes they cover~\citep{allen2023assessingcalibrationmultivariateprobabilistic} (Appendix~\ref{sec:app_eval}). In calibration, the MDN and state tokenization were overconfident in orientation in every condition, which is consistent with mixture density networks collapsing onto sharp modes~\citep{makansi2019overcoming} and with state tokenization predicting only token centers. The two metrics thus complement the Brier score by checking whether the predicted spread is accurate, which exposes the overconfidence above.

\begin{figure}[!t]
  \centering
  \includegraphics[width=\columnwidth]{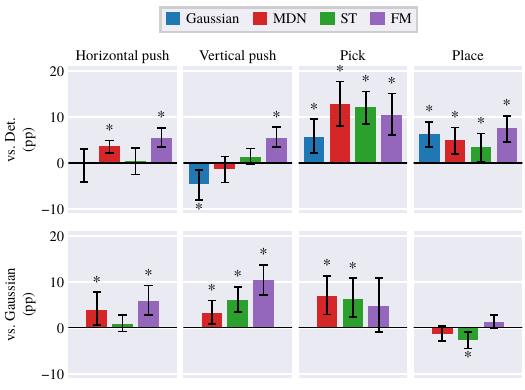}
  \caption{Paired difference in simulated single-step task success (percentage points) at $\epsilon_{\mathrm{pos}}=5$\,cm and $\epsilon_{\mathrm{rot}}=20^\circ$. Error bars show 95\% object-level bootstrap intervals; $*$ marks intervals that exclude zero.}
  \label{fig:success_diff}
\end{figure}

\subsection{Simulation Planning Rollouts}

\begin{figure}[t]
  \centering
  \includegraphics[width=\columnwidth]{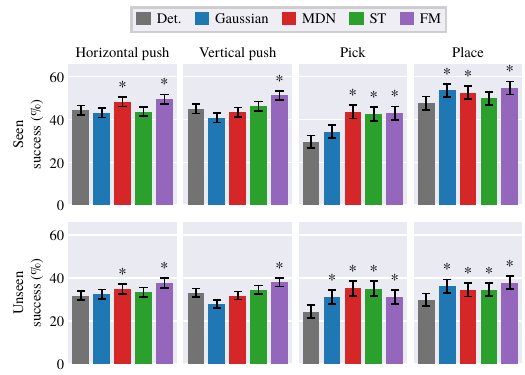}
  \caption{Simulated single-step task success rate on seen and unseen objects at $\epsilon_{\mathrm{pos}}=5$\,cm and $\epsilon_{\mathrm{rot}}=20^\circ$. Error bars show 95\% Wilson intervals; $*$ marks models significantly higher than the deterministic model.}
  \label{fig:success_seen_unseen}
\end{figure}

\paragraph{Single-step Planning}
Flow matching was the only model that significantly improved the success rate over the deterministic model on every skill (Figure~\ref{fig:success_diff}). This advantage held on unseen objects, although success dropped for every model (Figure~\ref{fig:success_seen_unseen}). For reference, replaying the demonstrated action succeeded on 79.7\% and 83.7\% of the pushes (Appendix~\ref{sec:app_success}).

Every probabilistic model significantly outperformed the deterministic model on pick and place. On push, however, the benefit was less consistent. Only the MDN on horizontal push and flow matching on both pushes improved significantly, and the Gaussian was significantly worse on vertical push. A predicted distribution helps planning beyond its prediction accuracy. The deterministic model made the most plausible predictions on place yet the least successful plans, and on pick, the Gaussian planned better than it despite a worse Brier score. This is consistent with the planning cost. The expected cost averages the goal negative log-likelihood over sampled outcomes~\citep{22-arxiv-uncertaintygoalplanning}, so it penalizes actions whose outcomes spread widely, and any predicted distribution steers the planner toward actions that reach the goal reliably.

All multimodal models succeeded more often than the Gaussian on push and pick, although only the MDN did so significantly on all three skills. On place, no multimodal model significantly outperformed the Gaussian, and state tokenization was significantly worse. This is consistent with the plausibility results, where the Gaussian predicted floating outcomes less often on place than the multimodal models (Figure~\ref{fig:prediction_quality}, top).

\paragraph{Multi-step Planning}

\begin{figure*}[t]
\centering
\includegraphics[width=0.8\textwidth]{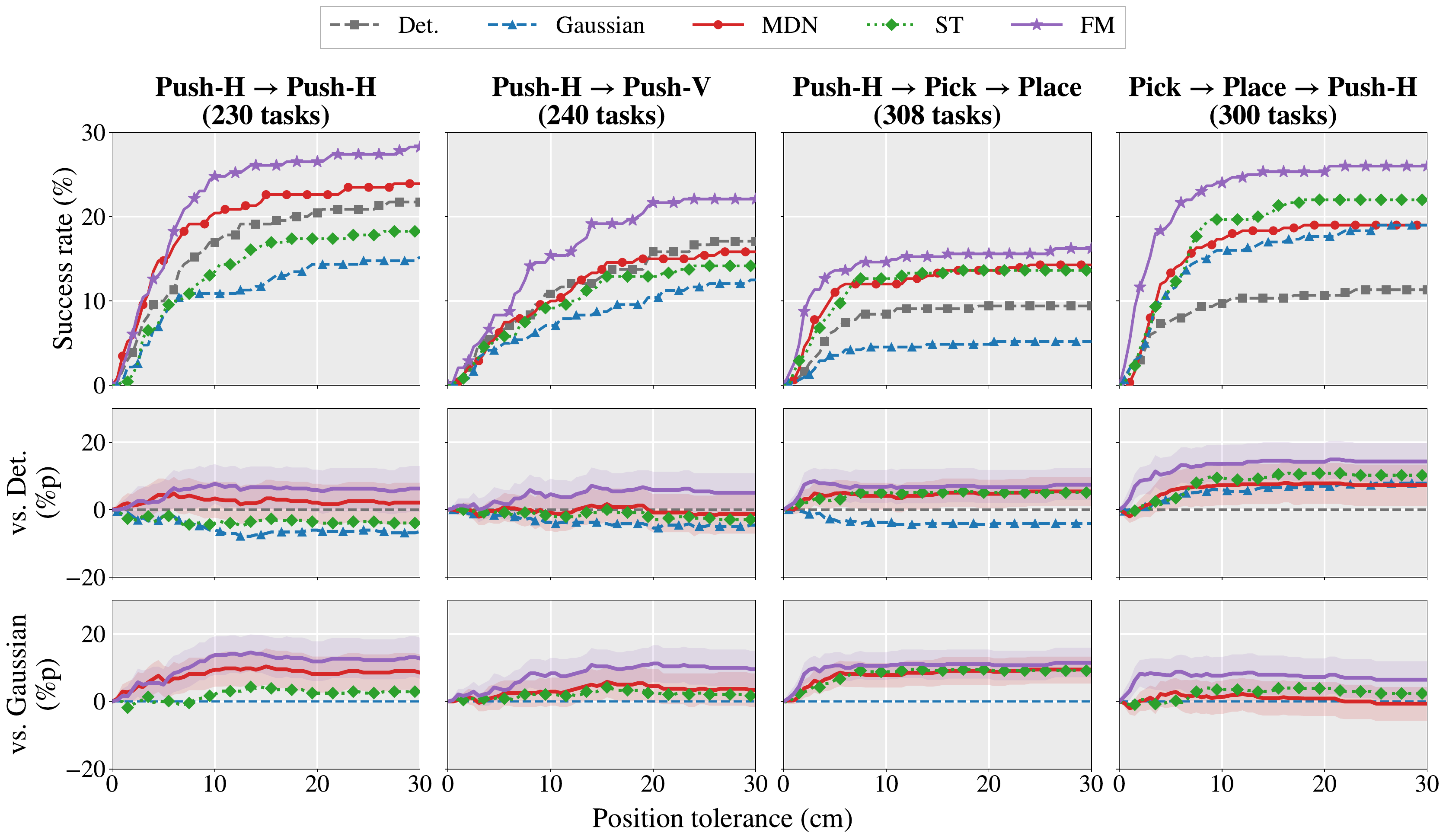}
\caption{Multi-step success over the position tolerance. For evaluation, we fixed $\epsilon_{\mathrm{rot}}=20^\circ$ and varied the position tolerance. Each model plans the skeleton jointly and executes it open-loop, on tasks shared by all models. Top: success rate. Middle and bottom: paired difference from the deterministic model and from the Gaussian. Shaded bands show 95\% bootstrap intervals for the MDN and flow matching, resampling tasks within each object with objects weighted equally.}
\label{fig:multistep_cdf}
\end{figure*}

\begin{figure*}[t]
  \centering
  \includegraphics[width=\textwidth]{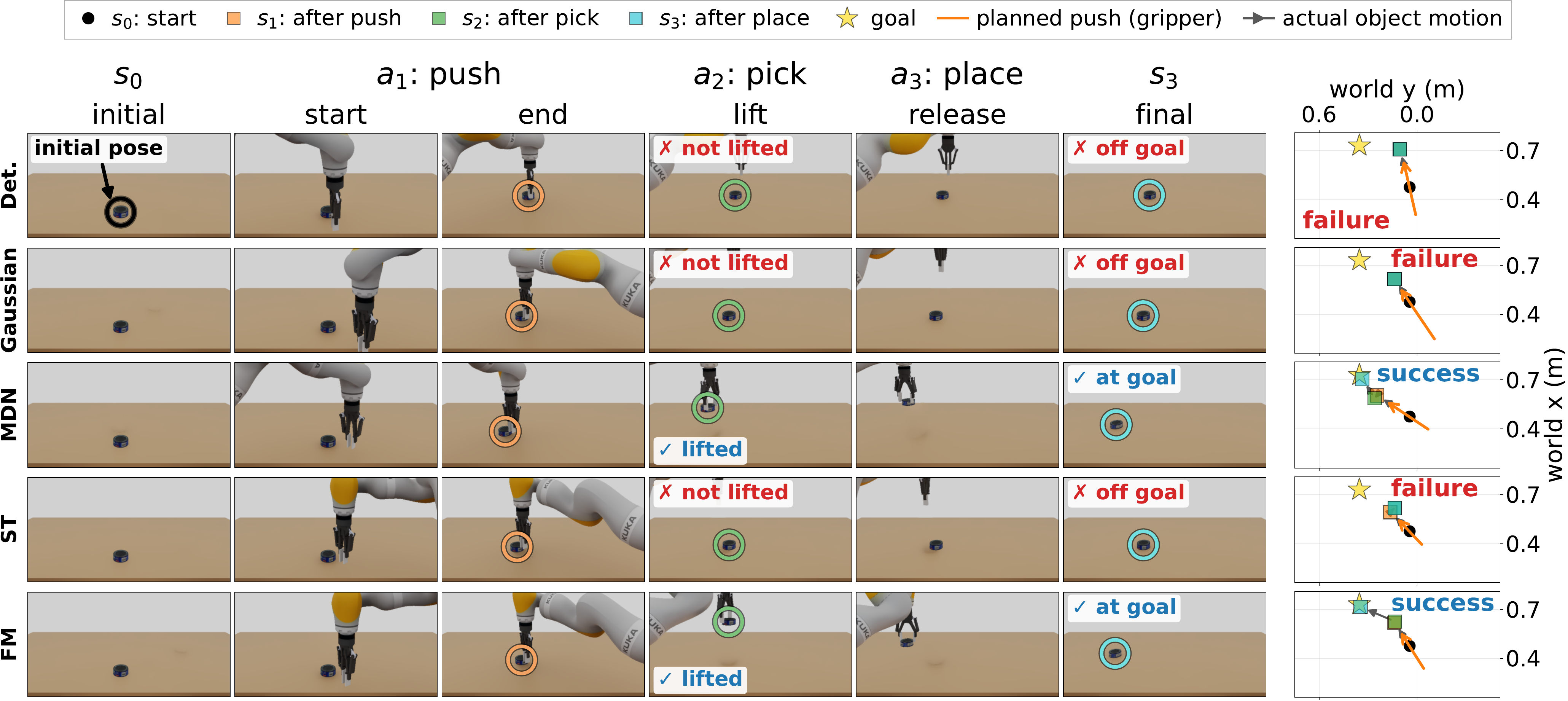}
  \caption{Simulation rollouts of the Push-H$\to$Pick$\to$Place skeleton for one task (tuna can), each planned by a different model from the same start and goal.
  ST and FM denote state tokenization and flow matching. The rightmost column shows the top view of the object motion, starting from $s_0$ (black dot), with the planned push (orange line), the actual object motion (gray arrows), the object poses $s_1$, $s_2$, and $s_3$ after the push, pick, and place (squares), and the goal (star). In this example, the grasp planned after the push does not lift the can with the deterministic model, the Gaussian, and state tokenization, so the can stays where the push left it, whereas the MDN and flow matching lift the can and place it at the goal. This episode illustrates a failure mode and does not represent aggregate performance on this skeleton.}
  \label{fig:multistep_rollout}
\end{figure*}

Figure~\ref{fig:multistep_cdf} shows multi-step success as a function of the position tolerance, and Figure~\ref{fig:multistep_rollout} shows an example rollout. Each model planned the whole plan skeleton jointly and executed it open-loop. Because each subsequent skill was planned on the predicted outcome of the previous one, prediction errors carried through the steps to the final pose. This setting tests whether the predicted distributions remain useful over longer horizons. For evaluation, we fixed $\epsilon_{\mathrm{rot}}=20^\circ$ and varied the position tolerance. We evaluated two two-step skeletons (Push-H$\to$Push-H and Push-H$\to$Push-V) and two three-step skeletons (Push-H$\to$Pick$\to$Place and Pick$\to$Place$\to$Push-H), where H and V denote horizontal and vertical push.

Flow matching achieved the highest success rate on all four plan skeletons at nearly every tolerance (Figure~\ref{fig:multistep_cdf}). The MDN outperformed the deterministic model on three of the four plan skeletons. State tokenization helped only once a pick was involved and fell below the deterministic model on the push-only skeletons, in line with its low plausibility on horizontal push. The Gaussian fell below the deterministic model on the three skeletons that begin with a horizontal push and exceeded it only on Pick$\to$Place$\to$Push-H. Push-H$\to$Pick$\to$Place was the hardest skeleton, which is consistent with the pick being planned on a predicted pose, where a small pose error makes the grasp fail. The advantage of flow matching and the MDN over the deterministic model is consistent with how the parameters of the earlier skills are chosen. The deterministic model commits to a single outcome, whereas the probabilistic models score each candidate over its sampled outcomes and so favor parameters whose outcomes all leave the next skill feasible.

Overall, flow matching outperformed both baselines on every skeleton, and the MDN outperformed each of them on three of the four. The Gaussian, in contrast, helped on single-step pick and place yet fell below the deterministic model on three of the four skeletons. This suggests that representing multiple outcomes, rather than uncertainty alone, is what helps when predictions are chained.

\subsection{Real-world Planning Rollouts}

\paragraph{Single-step Planning}
We evaluated every model on every skill on a real robot with five objects. Three were seen in training (cracker box, mug, and mustard bottle) and two were unseen (a penguin toy, and a screwdriver for vertical push or a wipes container for horizontal push). We used three start-goal pairs per object, giving 15 tasks per model. 
We estimated object poses from a single RGB-D view and checked every result against the camera video (Appendix~\ref{sec:app_success}). 

\begin{figure}[t]
  \centering
  \includegraphics[width=\columnwidth]{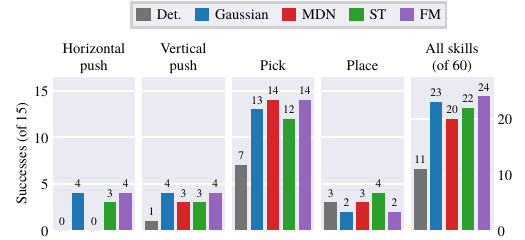}
  \caption{Real-robot single-step task success. Left: successful tasks out of 15 per skill. Right: total over the four skills (out of 60).}
  \label{fig:real_success}
\end{figure}

Figure~\ref{fig:real_success} reports the results on the real robot. On pick, every probabilistic model succeeded about twice as often as the deterministic model, in line with the pick results in simulation. The deterministic model was also overconfident. It predicted success on 12 of its 15 chosen picks but succeeded on 7.

\paragraph{Multi-step Planning}

As a qualitative example, Figure~\ref{fig:real_multistep} shows one run per model of Push-H$\to$Pick$\to$Place, the hardest skeleton in simulation, on the seen potted meat can from the same start and goal. The Gaussian failed at the push step, and the deterministic model failed at the pick step. The MDN and state tokenization completed every step but left the object on its side. Only flow matching reached the goal, in line with its highest success on every skeleton in simulation.

\section{Limitations and Future Work}
The advantage of the multimodal models in predicting which outcome occurs did not carry over to planning uniformly. In single-step planning, they improved less over the Gaussian than over the deterministic model, and on the real robot, success on push and place was low for every model.

Our multi-step plans are executed open-loop so that we can evaluate the effects of error propagation through the plans. A natural next step is to replan from the observed outcome after each skill, where a multimodal model could prepare a plan for each outcome it predicts. State tokenization is trained with a cross-entropy loss that penalizes a neighboring token, which encodes a nearby pose, as much as a distant one. A loss that accounts for the distance between tokens, such as the earth mover's distance~\citep{frogner2015learningwassersteinloss}, could credit near misses and improve its plausibility and planning. Our simulation fixes the friction and mass of each object, so the learned distributions do not account for uncertainty in these physical properties. Randomizing them during data collection would let a probabilistic SEM capture this source of variation as well. Finally, we model only geometric effects on a single target object, and we plan to extend probabilistic SEMs to multiple interacting objects and to semantic or relational effects.

\section{Conclusion}
\label{sec:conclusion}

We studied probabilistic skill-effect models that represent multiple outcomes of contact-rich skills from partial point clouds. Comparing mixture density networks, state tokenization, and flow matching with deterministic and Gaussian baselines on push, pick, and place with seen and unseen objects, we found that the MDN and flow matching anticipate which outcome occurs more accurately and plan more consistently, with flow matching the most reliable. In contrast, the deterministic model is overconfident, and the single Gaussian struggles where effects split, as in pick, and falls below the deterministic model on most multi-step skeletons. These results suggest that multimodal skill-effect models are useful for planning because they capture distinct modes rather than uncertainty alone.

\section*{Acknowledgment}
This work was supported by DARPA under grant HR0011-24-9-0423. The authors used Claude (Anthropic) for language editing and assistance with animations, visual layout, and editing of the supplementary video. All experimental data and results were produced by the authors.

\bibliography{ref}
\bibliographystyle{IEEEtranN}

\newpage
\appendix

\begin{figure}[H]
\centering
\includegraphics[width=0.8\columnwidth]{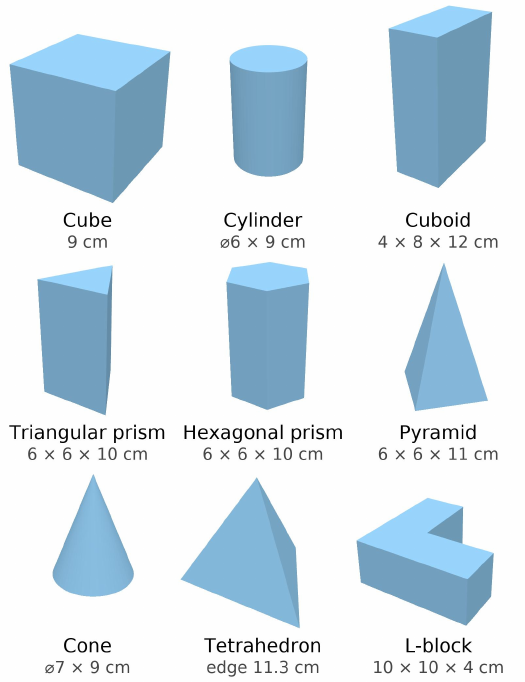}
\caption{One representative of each of the nine primitive families, rendered at the same scale with nominal dimensions.}
\label{fig:primitives}
\end{figure}

\subsection{Training Details}
\label{sec:app_training}

\textbf{Deterministic baseline:} The deterministic baseline follows Points2Plans~\citep{24-arxiv-points2plans}, which predicts only the position, and we add a rotation term. The decoder predicts a single relative pose \(\widehat{\delta p}_t=\Dec_{\mathrm{det}}(\delta\mathbf{z}_t)=(\widehat{\delta \mathbf{t}}_t,\widehat{\delta R}_t)\), where \(\delta\mathbf{z}_t\) is the predicted latent update for the target object. It is trained with
\(
\mathcal{L}_{\mathrm{det}}
=
w_{\mathrm{pos}}\sqrt{c_{\mathrm{pos}}\,\ell_{\mathrm{pos}}}
+
w_{\mathrm{rot}}\,\ell_{\mathrm{rot}},
\)
where \(\ell_{\mathrm{pos}}\) is the L1 norm of the translation error \(\delta \mathbf{t}_t-\widehat{\delta \mathbf{t}}_t\) and \(\ell_{\mathrm{rot}}\) is the squared geodesic distance \(\lVert\mathrm{Log}(\delta R_t^\top\widehat{\delta R}_t)\rVert_2^2\), both averaged over the batch. We use \(w_{\mathrm{pos}}=3\), \(c_{\mathrm{pos}}=5\), and \(w_{\mathrm{rot}}=2\). As in Points2Plans, it is also trained with the latent transition loss \(\mathcal{L}_{\mathrm{reg}}\), the mean squared error between \(\mathbf{z}_t+\delta\mathbf{z}_t\) and \(\Enc_s(S_{t+1})\), and minimizes \(\lambda_r\mathcal{L}_{\mathrm{reg}}+\lambda_{\mathrm{task}}\mathcal{L}_{\mathrm{det}}\) with \(\lambda_r=2\) and \(\lambda_{\mathrm{task}}=1\).

\begin{table}[t]
\centering
\caption{Test objects and episodes per skill.}
\label{tab:test_objects}
\footnotesize
\setlength{\tabcolsep}{2pt}
\begin{tabular}{lrrrr}
\toprule
Object & \makecell{Horiz.\\push} & \makecell{Vert.\\push} & Place & Pick \\
\midrule
\rowcolor{rowalt}\multicolumn{5}{l}{\textbf{Seen objects} (held-out episodes)} \\
Toy airplane                    & 205 & 196 & 143 & 401 \\
Cracker box                     & 202 & 202 & 131 & 399 \\
Cube                            & 199 & 200 & 183 & 398 \\
Hexagonal prism (tall)          & 192 & 202 & 177 & 399 \\
Mug                             & 192 & 202 & 150 & 398 \\
Mustard bottle                  & 195 & 203 & 122 & 400 \\
Potted meat can                 & 196 & 205 & 171 & 399 \\
Sugar box                       & 198 & 205 & 162 & 400 \\
Tuna fish can                   & 195 & 203 & 157 & 400 \\
Triangular prism                & 189 &  -- & 174 & 400 \\
L-block                         &  -- & 205 & 179 & 400 \\
\textit{Subtotal (objects)}     & \textit{1,963 (10)} & \textit{2,023 (10)} & \textit{1,749 (11)} & \textit{4,394 (11)} \\
\midrule
\rowcolor{rowalt}\multicolumn{5}{l}{\textbf{Unseen objects}} \\
Nestl\'e Crunch cookie box      & 201 & 204 & 151 & 400 \\
Room Essentials mug             & 191 & 202 & 130 & 397 \\
Stacking bear                   & 197 & 203 & 157 & 400 \\
Threshold porcelain pitcher     & 202 & 200 & 170 & 398 \\
Threshold porcelain ramekin     & 194 & 202 & 115 & 398 \\
Twinlab Nitric Fuel bottle      & 191 & 203 & 157 & 400 \\
U by Kotex box                  & 194 & 203 & 181 & 399 \\
Vans cereal box                 & 203 & 202 & 158 & 400 \\
Weston sausage tonic bottle     & 200 & 202 & 136 & 399 \\
Schleich Hereford bull          &  -- & 204 & 165 & 400 \\
Ecoforms plant saucer (SQ8)     &  -- & 202 &  -- &  -- \\
Ecoforms plant saucer (S14)     &  -- &  59 &  -- &  -- \\
\textit{Subtotal (objects)}     & \textit{1,773 (9)} & \textit{2,286 (12)} & \textit{1,520 (10)} & \textit{3,991 (10)} \\
\midrule
\textbf{Total (objects)}        & \textbf{3,736 (19)} & \textbf{4,309 (22)} & \textbf{3,269 (21)} & \textbf{8,385 (21)} \\
\bottomrule
\end{tabular}
\end{table}

\textbf{Training parameters:} We train all models for 600 epochs with batch size 512 in fp32, with one Adam optimizer each for the encoder, the latent dynamics model, and the decoder. The learning rate is \(3\times10^{-4}\) (\(1.5\times10^{-4}\) for the encoder of the state-tokenization model), warmed up linearly over the first 500 steps and constant afterward, and gradients are clipped to norm 1.0. Each input observation consists of partial point clouds of the target object and the table, each downsampled to at most 512 points via farthest point sampling. The task loss is applied to the target object only, with the table as context. The probabilistic models are trained without \(\mathcal{L}_{\mathrm{reg}}\) (Appendix~\ref{sec:app_reg}). For state tokenization, the no-motion token, in which the target object stays at rest, dominates the data. To prevent the model from collapsing to this token, each training example is drawn from the no-motion examples with probability 0.2 and from the remaining examples with probability 0.8. Validation uses the original distribution.

\textbf{Training and evaluation datasets:} Of the 94 training objects (Section~\ref{sec:datasets}), horizontal push uses 57, vertical push uses all 94, and pick and place use 91. The 50 primitives span nine families (Figure~\ref{fig:primitives}), namely 6 cubes, 14 cylinders, 18 cuboids, 3 triangular and 3 hexagonal prisms, 2 pyramids, a cone, a tetrahedron, and 2 L-blocks. Their longest side is between 5 and 18\,cm and their mass is 0.5\,kg. They include shapes with distinct contact behavior, such as flat discs as thin as 2.5\,cm, thin rods 10--12\,cm tall on a 3--4\,cm base, and non-convex L-blocks whose center of mass is offset from their geometric center. Friction and mass are fixed for each object.

For push, each nominal execution has five copies with the object start position shifted by Gaussian noise (2\,cm standard deviation in $x$ and $y$) and the yaw by uniform noise within $\pm30^\circ$. For pick, uniform noise within $\pm2$\,cm per axis and $\pm45^\circ$ in yaw is added to the grasp pose, scaled by 0.3 with probability 0.5, and the lift height is sampled from 0.1--0.5\,m.
For place, the release yaw is uniform within $\pm180^\circ$, and the release height is sampled from 0--0.1\,m with probability 0.2 and from 0.1--0.5\,m otherwise.
A push example counts as a failed execution if the object moves less than 2\,cm. Two-thirds of the failed pick examples are hard no-ops, in which the gripper contacts the object but fails to lift it.

We recorded the executed action and the partial-view point clouds before and after each execution.
Of the episodes on the training objects, a random 20\% was held out for validation (13{,}022, 20{,}626, 25{,}952, and 12{,}225 for horizontal push, vertical push, pick, and place), leaving 52{,}084, 82{,}502, 103{,}804, and 48{,}899 episodes for training. Table~\ref{tab:test_objects} lists the test objects of each skill.

\begin{figure*}[t]
  \centering
  \includegraphics[width=0.7\textwidth]{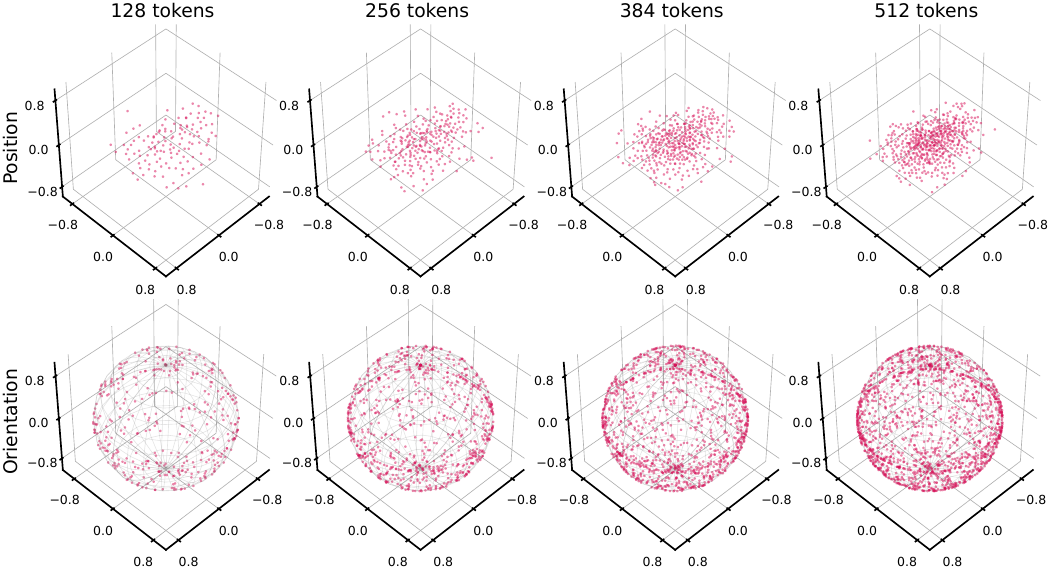}
  \caption{State-token sets with 128, 256, 384, and 512 tokens. Top: translation of each token. Bottom: rotation of each token, shown as its rotated coordinate axes. Larger sets cover the same range more densely, and translation tokens concentrate around small motions.}
  \label{fig:token_sets}
\end{figure*}

\begin{table*}[t]
\centering
\caption{Prediction metrics across three training seeds.}
\label{tab:seed_eval}
\begin{tabular}{llccccc}
\toprule
Skill & Metric & Det. & Gaussian & MDN & ST & FM \\
\midrule
\rowcolor{rowalt}Horizontal push & Brier score & 0.748 $\pm$ 0.081 & 0.674 $\pm$ 0.037 & 0.477 $\pm$ 0.015 & 0.479 $\pm$ 0.011 & \textbf{0.400 $\pm$ 0.001} \\
 & Plausibility (\%) & 57.0 $\pm$ 4.8 & 51.3 $\pm$ 7.1 & 65.0 $\pm$ 4.2 & 43.6 $\pm$ 0.3 & \textbf{68.8 $\pm$ 1.8} \\
 & ES position (cm) & 8.20 $\pm$ 0.10 & 6.63 $\pm$ 0.26 & 6.16 $\pm$ 0.08 & 6.64 $\pm$ 0.10 & \textbf{5.55 $\pm$ 0.04} \\
 & ES orientation & 0.976 $\pm$ 0.028 & 0.716 $\pm$ 0.018 & 0.681 $\pm$ 0.022 & 0.704 $\pm$ 0.018 & \textbf{0.641 $\pm$ 0.009} \\
 & Calib.\ err.\ position & -- & 0.187 $\pm$ 0.039 & 0.179 $\pm$ 0.032 & 0.198 $\pm$ 0.039 & \textbf{0.132 $\pm$ 0.014} \\
 & Calib.\ err.\ orientation & -- & 0.186 $\pm$ 0.040 & 0.192 $\pm$ 0.037 & 0.286 $\pm$ 0.023 & \textbf{0.177 $\pm$ 0.016} \\
\midrule
\rowcolor{rowalt}Vertical push & Brier score & 0.575 $\pm$ 0.041 & 0.702 $\pm$ 0.016 & 0.386 $\pm$ 0.020 & \textbf{0.326 $\pm$ 0.012} & 0.328 $\pm$ 0.004 \\
 & Plausibility (\%) & 69.9 $\pm$ 1.9 & 49.8 $\pm$ 0.8 & 73.4 $\pm$ 1.4 & 62.5 $\pm$ 1.4 & \textbf{78.0 $\pm$ 0.5} \\
 & ES position (cm) & 10.06 $\pm$ 0.01 & 7.97 $\pm$ 0.35 & 7.73 $\pm$ 0.25 & 7.34 $\pm$ 0.09 & \textbf{7.13 $\pm$ 0.21} \\
 & ES orientation & 1.042 $\pm$ 0.020 & 0.730 $\pm$ 0.013 & 0.733 $\pm$ 0.010 & 0.733 $\pm$ 0.003 & \textbf{0.681 $\pm$ 0.016} \\
 & Calib.\ err.\ position & -- & \textbf{0.054 $\pm$ 0.021} & 0.152 $\pm$ 0.060 & 0.210 $\pm$ 0.007 & 0.142 $\pm$ 0.024 \\
 & Calib.\ err.\ orientation & -- & \textbf{0.073 $\pm$ 0.006} & 0.198 $\pm$ 0.059 & 0.259 $\pm$ 0.007 & 0.150 $\pm$ 0.021 \\
\midrule
\rowcolor{rowalt}Pick (all) & Brier score & 0.724 $\pm$ 0.074 & 0.934 $\pm$ 0.115 & 0.514 $\pm$ 0.009 & 0.619 $\pm$ 0.004 & \textbf{0.454 $\pm$ 0.006} \\
\addlinespace[2pt]
Pick (nominal) & Plausibility (\%) & 84.9 $\pm$ 2.0 & 57.4 $\pm$ 7.3 & 82.5 $\pm$ 0.9 & 78.1 $\pm$ 0.5 & \textbf{86.5 $\pm$ 0.9} \\
 & ES position (cm) & 6.39 $\pm$ 0.26 & 5.30 $\pm$ 0.48 & 5.22 $\pm$ 0.26 & \textbf{4.99 $\pm$ 0.25} & 5.16 $\pm$ 0.15 \\
 & ES orientation & 0.533 $\pm$ 0.016 & 0.405 $\pm$ 0.017 & 0.353 $\pm$ 0.005 & 0.387 $\pm$ 0.006 & \textbf{0.348 $\pm$ 0.012} \\
 & Calib.\ err.\ position & -- & 0.077 $\pm$ 0.040 & \textbf{0.038 $\pm$ 0.018} & 0.092 $\pm$ 0.059 & 0.087 $\pm$ 0.023 \\
 & Calib.\ err.\ orientation & -- & 0.096 $\pm$ 0.045 & 0.095 $\pm$ 0.031 & 0.204 $\pm$ 0.034 & \textbf{0.083 $\pm$ 0.025} \\
\addlinespace[2pt]
Pick (hard no-op) & Plausibility (\%) & 62.7 $\pm$ 5.4 & 26.8 $\pm$ 9.0 & 74.3 $\pm$ 1.5 & 69.7 $\pm$ 2.6 & \textbf{79.9 $\pm$ 1.3} \\
 & ES position (cm) & 9.07 $\pm$ 0.81 & \textbf{6.77 $\pm$ 0.56} & 8.40 $\pm$ 0.33 & 9.61 $\pm$ 0.02 & 7.91 $\pm$ 0.12 \\
 & ES orientation & 1.042 $\pm$ 0.013 & \textbf{0.739 $\pm$ 0.010} & 0.772 $\pm$ 0.011 & 0.805 $\pm$ 0.011 & 0.757 $\pm$ 0.015 \\
 & Calib.\ err.\ position & -- & \textbf{0.071 $\pm$ 0.019} & 0.150 $\pm$ 0.031 & 0.213 $\pm$ 0.034 & 0.182 $\pm$ 0.019 \\
 & Calib.\ err.\ orientation & -- & \textbf{0.087 $\pm$ 0.012} & 0.190 $\pm$ 0.035 & 0.261 $\pm$ 0.024 & 0.172 $\pm$ 0.021 \\
\midrule
\rowcolor{rowalt}Place & Brier score & 1.034 $\pm$ 0.004 & 1.124 $\pm$ 0.117 & 0.728 $\pm$ 0.023 & 0.676 $\pm$ 0.018 & \textbf{0.588 $\pm$ 0.011} \\
 & Plausibility (\%) & \textbf{39.4 $\pm$ 3.6} & 18.0 $\pm$ 3.9 & 30.8 $\pm$ 2.0 & 21.9 $\pm$ 0.7 & 27.5 $\pm$ 0.5 \\
 & ES position (cm) & 7.83 $\pm$ 0.06 & 5.60 $\pm$ 0.11 & \textbf{5.49 $\pm$ 0.22} & 5.82 $\pm$ 0.04 & 5.51 $\pm$ 0.06 \\
 & ES orientation & 1.100 $\pm$ 0.005 & 0.728 $\pm$ 0.010 & 0.714 $\pm$ 0.030 & 0.744 $\pm$ 0.003 & \textbf{0.689 $\pm$ 0.007} \\
 & Calib.\ err.\ position & -- & 0.150 $\pm$ 0.063 & \textbf{0.107 $\pm$ 0.059} & 0.192 $\pm$ 0.015 & 0.148 $\pm$ 0.014 \\
 & Calib.\ err.\ orientation & -- & 0.091 $\pm$ 0.011 & \textbf{0.076 $\pm$ 0.042} & 0.204 $\pm$ 0.015 & 0.086 $\pm$ 0.009 \\
\bottomrule
\end{tabular}\\[2pt]
\parbox{\textwidth}{\footnotesize Mean and standard deviation over three training seeds. Shaded rows show the Brier score, the metric discussed in the main text. Bold marks the best mean in each row. ES is the energy score. The deterministic model has no calibration error because it predicts a single outcome.}
\end{table*}

\subsection{Evaluation Protocol}
\label{sec:app_eval}

\textbf{Outcome-class Brier score:}
For an episode, let $\hat p_c$ be the fraction of the $M$ predicted samples assigned to class $c$, and let $o_c=1$ if the observed outcome belongs to class $c$ and $0$ otherwise. The score is $\sum_c(\hat p_c-o_c)^2$, averaged over episodes. The deterministic model has $\hat p_c\in\{0,1\}$. For push and place, an object whose tilt is within $10^\circ$ of its reference rest pose is \emph{keep}, within $10^\circ$ of another rest pose is \emph{topple}, and otherwise \emph{off}. For push, the reference is the rest pose the object starts in. For place, it is the rest pose the object is held in at release. Rest poses that are identical under the object's symmetries count as one.
For pick, an object is \emph{held} if it lies between the gripper's fingertips, within a 2\,cm margin. Otherwise, it is \emph{ungrasped} if its lowest point lies within 1\,cm of the table, and \emph{off} if not.

\textbf{Physical plausibility:}
A predicted outcome is plausible if the lowest point of the object's convex hull lies within 1\,cm of the table surface and the object's tilt is within $10^\circ$ of one of its rest poses. For pick, an object held by the gripper is exempt from the floating and tilt checks but must still not penetrate the table. Object meshes are used only for evaluation, and the ground truth is at least 97.9\% plausible on every skill.

\textbf{Energy score:}
The energy score is the multivariate generalization of the continuous ranked probability score (CRPS)~\citep{gneiting2007strictly,szekely2013energy} and has recently been advocated for evaluating multimodal trajectory predictions~\citep{shahroudi2024energyscore}. Given the predictive distribution $P(\cdot\mid \mathbf{z}_t,A_t)$, a single observed outcome $\delta p_t$, and a distance $d(\cdot,\cdot)$, the energy score is
\begin{equation}
\mathrm{ES}_d\!\left(P,\delta p_t\right)
=
\mathbb{E}\!\left[d(X,\delta p_t)\right]
-\tfrac{1}{2}\,
\mathbb{E}\!\left[d(X,X')\right],
\label{eq:es_def}
\end{equation}
where $X,X'\sim P(\cdot\mid \mathbf{z}_t,A_t)$ are independent samples.
The first term is the expected distance to the observed outcome, and the second credits the spread of the predicted distribution.
The energy score is a strictly proper scoring rule~\citep{gneiting2007strictly}, but in practice it is only weakly sensitive to the number of modes~\citep{marcotte2023regions}.

We estimate it from $M$ samples $\{\widehat{\delta p}^{(m)}\}_{m=1}^{M}\sim P$ as
\begin{equation}
\begin{aligned}
\widehat{\mathrm{ES}}_d
={}&
\frac{1}{M}\sum_{m=1}^{M}
d\big(\widehat{\delta p}^{(m)},\delta p_t\big)\\
&-
\frac{1}{2M(M{-}1)}\sum_{m\neq m'}
d\big(\widehat{\delta p}^{(m)},\widehat{\delta p}^{(m')}\big),
\end{aligned}
\label{eq:es_estimator}
\end{equation}
using the unbiased estimator of the dispersion term.
For the deterministic model, the second term vanishes and the score reduces to the prediction error. 
Because position and orientation have different units, the position score $\mathrm{ES}_{\mathrm{pos}}$ uses $d_{\mathrm{pos}}(\delta\mathbf{t},\delta\mathbf{t}')=\|\delta\mathbf{t}-\delta\mathbf{t}'\|_2$, and the orientation score $\mathrm{ES}_{\mathrm{rot}}$ uses the chordal distance $d_{\mathrm{rot}}(\delta R,\delta R')=\|\delta R-\delta R'\|_F$, the Euclidean distance between rotation matrices~\citep{huynh2009metrics,hartley2013rotation}.
The chordal distance relates to the geodesic angle $\theta$ between two rotations by $d_{\mathrm{rot}}=2\sqrt{2}\sin(\theta/2)$, so a reported score can be read as an angle.
Training uses the geodesic distance instead (Section~\ref{sec:training}). We use $M{=}100$ for all models and report scores averaged over the test set.

\begin{table}[t]
\centering
\caption{Output-head hyperparameter selection.}
\label{tab:head_sweep}
\footnotesize\setlength{\tabcolsep}{2pt}
\begin{tabular}{lllccccccc}
\toprule
Model & \multicolumn{2}{c}{Setting} & \multicolumn{2}{c}{ES\,$\downarrow$} & \multicolumn{2}{c}{Calib.\ err.\,$\downarrow$} & \makecell{Plaus.\\(\%)\,$\uparrow$} & \multicolumn{2}{c}{Time (ms)\,$\downarrow$} \\
\cmidrule(lr){4-5}\cmidrule(lr){6-7}\cmidrule(lr){9-10}
 & & & pos (cm) & rot & pos & rot & & $b=1$ & $b=64$ \\
\midrule
\multirow{4}{*}{MDN} & \multirow{4}{*}{$K$} & 1 & 3.66 & 0.432 & 0.185 & \textbf{0.185} & 70.8 & 26.6 & 1.59 \\
                     &                      & 3 & \textbf{3.26} & 0.420 & 0.201 & 0.224 & 81.0 & 28.4 & 1.62 \\
                     &                      & 5$^\star$ & \textbf{3.26} & \textbf{0.416} & \textbf{0.176} & 0.228 & 81.5 & 29.2 & 1.67 \\
                     &                      & 7 & 3.34 & 0.425 & 0.194 & 0.232 & \textbf{82.1} & 30.7 & 1.70 \\
\midrule
\multirow{4}{*}{ST} & \multirow{4}{*}{$|\mathcal{V}|$} & 128 & 5.62 & 0.464 & 0.275 & 0.333 & 31.3 & 25.6 & 1.97 \\
                    &                                  & 256 & 4.77 & 0.439 & 0.234 & 0.315 & 53.7 & 26.1 & 2.06 \\
                    &                                  & 384 & 3.99 & 0.437 & 0.206 & 0.299 & 57.2 & 27.3 & 2.15 \\
                    &                                  & 512$^\star$ & \textbf{3.89} & \textbf{0.427} & \textbf{0.197} & \textbf{0.289} & \textbf{59.2} & 26.4 & 2.21 \\
\midrule
\multirow{8}{*}{FM} & \multirow{3}{*}{Size} & S & 3.06 & 0.380 & \textbf{0.060} & 0.070 & 85.2 & 46.7 & 2.70 \\
                    &                       & B$^\star$ & 3.05 & 0.370 & 0.061 & \textbf{0.050} & 86.2 & 53.7 & 4.77 \\
                    &                       & L & \textbf{2.97} & \textbf{0.366} & 0.072 & 0.062 & \textbf{88.1} & 69.9 & 19.37 \\
\cmidrule(lr){2-10}
                    & \multirow{5}{*}{Steps} & 1 & 3.32 & 0.429 & \textbf{0.053} & 0.060 & 49.0 & 32.1 & 2.49 \\
                    &                        & 3 & 3.05 & 0.372 & 0.070 & 0.066 & 83.8 & 43.1 & 3.63 \\
                    &                        & 5$^\star$ & 3.05 & 0.370 & 0.061 & \textbf{0.050} & \textbf{86.2} & 53.7 & 4.77 \\
                    &                        & 10 & \textbf{3.03} & \textbf{0.368} & 0.080 & 0.051 & 85.4 & 85.1 & 7.89 \\
                    &                        & 20 & \textbf{3.03} & \textbf{0.368} & 0.094 & 0.063 & 84.6 & 137.9 & 14.04 \\
\bottomrule
\end{tabular}\\[2pt]
\parbox{\columnwidth}{\footnotesize Horizontal-push validation set (13{,}022 episodes). $^\star$ marks the selected setting, and bold marks the best value in each sweep. ES is the energy score, $|\mathcal{V}|$ the number of tokens, and time is per episode at batch size $b$.}
\end{table}

\textbf{Calibration error:}
For an episode, let $x_1,\dots,x_M$ be the predicted samples and $x_{M+1}=y$ the observed outcome. Position and orientation use the distances $d_{\mathrm{pos}}$ and $d_{\mathrm{rot}}$ of the energy score, with no symmetry reduction. Each of the $M+1$ points receives the pre-rank
\begin{equation}
\eta_i=\frac{1}{M}\sum_{j\neq i}d(x_i,x_j),
\end{equation}
the mean distance from $x_i$ to the other points~\citep{knüppel2023scorebasedcalibrationtestingmultivariate,allen2023assessingcalibrationmultivariateprobabilistic}.
With
\(
n_<=\#\{i\le M:\eta_i<\eta_{M+1}\}
\)
and
\(
n_= =\#\{i\le M:\eta_i=\eta_{M+1}\},
\)
the outcome's randomized rank is
\begin{equation}
r=\frac{n_<+V\cdot(n_=+1)}{M+1},\qquad V\sim\mathcal{U}(0,1).
\end{equation}
If the model is calibrated, $r\sim\mathcal{U}(0,1)$. With $\hat F(\ell)$ the fraction of episodes with $r\le\ell$, the calibration error is
\begin{equation}
\mathrm{CE}=\frac{1}{99}\sum_{\ell\in\{0.01,\dots,0.99\}}\bigl|\hat F(\ell)-\ell\bigr|.
\end{equation}
$\hat F$ below the diagonal means the outcome lies outside the predicted spread too often (overconfident), and above it means the spread is too wide (underconfident). This pre-rank targets dispersion, so uniformity is necessary but not sufficient for multivariate calibration~\citep{allen2023assessingcalibrationmultivariateprobabilistic}.

\textbf{Task success:}
The planner receives the goal as a Gaussian~\citep{22-arxiv-uncertaintygoalplanning} over object poses with mean $(\mathbf{t}^\star,R^\star)$, and an execution succeeds if the final object pose $(\mathbf{t},R)$ satisfies
\begin{equation}
\begin{aligned}
e_{\mathrm{pos}} &= \lVert (\mathbf{t}-\mathbf{t}^\star)_{xy}\rVert \le \epsilon_{\mathrm{pos}},\\
e_{\mathrm{rot}} &= \min_{Q\in\mathcal{G}} \Delta_{\mathrm{yaw}}(RQ,\,R^\star) \le \epsilon_{\mathrm{rot}},\\
\gamma_{\mathrm{tilt}} &= \angle\big(R^{\top}e_z,\,{R^\star}^{\top}e_z\big) \le 10^\circ,
\end{aligned}
\label{eq:success}
\end{equation}
where $e_{\mathrm{pos}}$ and $e_{\mathrm{rot}}$ are the planar position error and the yaw error, and $\epsilon_{\mathrm{pos}}$ and $\epsilon_{\mathrm{rot}}$ are their tolerances. The term $\Delta_{\mathrm{yaw}}$ is the absolute yaw difference about the world up axis $e_z$, wrapped to $[0^\circ,180^\circ]$, and $\mathcal{G}$ is the object's symmetry group.
The tilt $\gamma_{\mathrm{tilt}}$ is the angle between the world up axis as seen from the final object frame and from the goal object frame. It is unaffected by yaw, so it tests whether the object rests on the same side as in the goal.
For pick, the object must also be lifted and held. A task also counts as failed if planning or execution fails. We report success over a grid of tolerances (Appendix~\ref{sec:app_success}). Success rates are reported with 95\% Wilson intervals. In simulation, we compare models by their paired differences in success over the same starts and goals, with 95\% object-level bootstrap intervals (2{,}000 resamples).

\subsection{Hyperparameters}
\label{sec:app_hyper}

We select each head's hyperparameters on the horizontal-push validation set and use the selected values for all skills. For the MDN, we vary the number of mixture components $K\in\{1,3,5,7\}$, and for state tokenization, the number of tokens $|\mathcal{V}|\in\{128, 256, 384, 512\}$. For flow matching, we sweep the size of the velocity network and the number of midpoint integration steps $\in\{1,3,5,10,20\}$. The sizes S, B, and L follow DiT~\citep{peebles_scalable_2023}, with depth and width 2/128, 4/256, and 8/512 and 1.08M, 4.58M, and 29.40M parameters. We evaluate the size with five steps and the number of steps with size B. Time is the mean over 100 runs after 10 warm-up runs on one NVIDIA RTX 6000 Ada GPU (fp32). Each call predicts $M{=}100$ samples for each of $b$ episodes from a fixed set of 200 validation episodes, and we report the time per episode for $b=1$ and $b=64$.

We weigh accuracy against inference time. For the MDN and state tokenization heads, the settings took nearly the same time, so we choose by the validation metrics. We take the lowest energy score and, among settings within 1\% of it, the highest physical plausibility.
For the MDN, $K{=}3$ and $K{=}5$ tied on the position energy score, and $K{=}5$ had the lowest orientation energy score and position calibration error. Adding more components did not lower either energy score, so we adopt $K{=}5$. For the state tokenization head, every metric improved with the number of tokens, with diminishing improvement toward 512, so we use 512 tokens. Larger sets cover the same range more densely, and the translation tokens concentrate around small motions (Figure~\ref{fig:token_sets}).
For flow matching, three to twenty integration steps tied on the energy score, and five steps gave the most plausible samples. A single step had the lowest position calibration error but only 49.0\% plausible samples, so calibration alone is not a sufficient criterion.
The network size is the one choice where time differed markedly. The large network lowered the position energy score by 2.6\% but was four times slower per episode at $b{=}64$, so we use size B with five integration steps. Because the larger network scored better on validation, this choice does not favor flow matching.

\begin{table}[t]
\centering
\caption{Encoder comparison.}
\label{tab:encoder_comparison}
\footnotesize\setlength{\tabcolsep}{2pt}
\resizebox{\columnwidth}{!}{%
\begin{tabular}{llccccccc}
\toprule
Head & Encoder & Params & \multicolumn{2}{c}{ES\,$\downarrow$} & \multicolumn{2}{c}{Calib.\ err.\,$\downarrow$} & Brier\,$\downarrow$ & \makecell{Plaus.\\(\%)\,$\uparrow$} \\
\cmidrule(lr){4-5}\cmidrule(lr){6-7}
 & & & pos (cm) & rot & pos & rot & & \\
\midrule
\multirow{3}{*}{FM}
 & PT-v3       & 38.8M & 5.55 & 0.644 & 0.130 & 0.181 & \textbf{0.400} & 70.7 \\
 & PT-v3-small & 1.18M & \textbf{5.53} & \textbf{0.643} & \textbf{0.126} & \textbf{0.173} & 0.402 & 69.7 \\
 & PointConv   & 1.16M & 5.88$^*$ & 0.670$^*$ & 0.136 & 0.179 & 0.414 & \textbf{73.0} \\
\midrule
\multirow{3}{*}{\makecell[l]{MDN}}
 & PT-v3       & 38.8M & 6.23 & 0.707 & 0.215 & 0.235 & 0.494 & \textbf{69.7} \\
 & PT-v3-small & 1.18M & \textbf{6.11} & \textbf{0.678} & 0.219 & 0.234 & \textbf{0.466} & 68.2 \\
 & PointConv   & 1.16M & 6.20 & 0.691 & \textbf{0.209} & \textbf{0.210}$^*$ & 0.502 & 61.7$^*$ \\
\bottomrule
\end{tabular}}\\[2pt]
\parbox{\columnwidth}{\footnotesize Horizontal-push test set (3{,}736 episodes, 19 objects). Bold marks the best value for each head, and $^*$ marks a significant paired difference from PT-v3. ES is the energy score.}
\end{table}

\subsection{Encoder Architecture Comparison}
\label{sec:app_encoder}

Table~\ref{tab:encoder_comparison} compares Point Transformer v3 (PT-v3)~\citep{Wuetal2024} with PointConv~\citep{19-cvpr-PointConv} as the point-cloud encoder on the horizontal-push test set (3{,}736 episodes, 19 objects), with the flow-matching and MDN heads. All rows use the same training protocol and the same latent dynamics model and decoder (17.3M parameters). The PT-v3 encoders take an object cloud of 128 to 512 points, and PointConv takes the same cloud padded to 512 points in preprocessing. Every encoder outputs a 256-dimensional feature, which is concatenated with the segment embedding into a 512-dimensional latent. PT-v3 is the full original encoder~\citep{Wuetal2024}, with five stages of serialized attention, channels 32/64/128/256/512, 2/2/2/6/2 blocks, and stride-2 grid pooling between stages (38.8M parameters). PointConv is a three-layer density-weighted set abstraction with 32/64/256 channels that reduces the cloud from 512 to 256, 64, and 1 point (1.16M parameters). To separate architecture from capacity, PT-v3-small keeps the PT-v3 block design and attention layout but uses four stages with channels 16/32/64/96 and 1/1/2/2 blocks, similar to the PointConv encoder in size (1.18M vs.\ 1.16M).

Encoder capacity did not matter here. PT-v3-small matched the full PT-v3 on every metric for both heads, with no significant paired difference despite 33 times fewer parameters. Compared with PT-v3, PointConv was significantly worse in energy score with the flow-matching head and in plausibility with the MDN head, although its orientation calibration error with the MDN head was significantly lower. No encoder differed significantly from PT-v3 in Brier score with either head, so the differences among encoders are smaller than the differences among heads in the main results (Figures~\ref{fig:brier} and~\ref{fig:prediction_quality}).

Since the encoders performed comparably, we choose PT-v3 because it processes point clouds of varying size natively. It encodes clouds of 128 to 512 points as they are, without the padding that PointConv requires.

\begin{table}[t]
\centering
\caption{Effect of the latent transition loss $\mathcal{L}_{\mathrm{reg}}$.}
\label{tab:reg}
\footnotesize\setlength{\tabcolsep}{2pt}
\resizebox{\columnwidth}{!}{%
\begin{tabular}{lccccccc}
\toprule
Head & \multicolumn{2}{c}{ES\,$\downarrow$} & \multicolumn{2}{c}{Calib.\ err.\,$\downarrow$} & Brier\,$\downarrow$ & \makecell{Plaus.\\(\%)\,$\uparrow$} & \makecell{Penetr.\\(\%)\,$\downarrow$} \\
\cmidrule(lr){2-3}\cmidrule(lr){4-5}
 & pos (cm) & rot & pos & rot & & & \\
\midrule
FM                                    & \textbf{5.55} & \textbf{0.644} & 0.130 & 0.181 & \textbf{0.400} & \textbf{70.7} & \textbf{23.5} \\
\quad$+\,\mathcal{L}_{\mathrm{reg}}$  & 8.27$^*$ & 0.888$^*$ & \textbf{0.110} & \textbf{0.159} & 0.761$^*$ & 61.4$^*$ & 32.3$^*$ \\
\midrule
MDN                           & \textbf{6.23} & \textbf{0.707} & 0.215 & 0.235 & \textbf{0.494} & \textbf{69.7} & \textbf{22.9} \\
\quad$+\,\mathcal{L}_{\mathrm{reg}}$  & 8.51$^*$ & 0.863$^*$ & \textbf{0.037}$^*$ & \textbf{0.054}$^*$ & 0.845$^*$ & 40.7$^*$ & 48.0$^*$ \\
\midrule
Gaussian                             & \textbf{6.87} & \textbf{0.732} & 0.213 & 0.203 & \textbf{0.675} & \textbf{55.6} & \textbf{30.7} \\
\quad$+\,\mathcal{L}_{\mathrm{reg}}$  & 8.12$^*$ & 0.870$^*$ & \textbf{0.044}$^*$ & \textbf{0.078}$^*$ & 1.311$^*$ & 25.6$^*$ & 48.1$^*$ \\
\midrule
ST                                    & 6.56 & 0.687 & 0.187 & 0.267 & 0.467 & 43.8 & \textbf{40.1} \\
\quad$+\,\mathcal{L}_{\mathrm{reg}}$  & \textbf{6.20}$^*$ & \textbf{0.678} & \textbf{0.174} & \textbf{0.239}$^*$ & \textbf{0.461} & 43.8 & 40.5 \\
\bottomrule
\end{tabular}}\\[2pt]
\parbox{\columnwidth}{\footnotesize Horizontal-push test set, with $\lambda_r=3$ for FM, MDN, and Gaussian and $\lambda_r=1$ for ST. Bold marks the better of the two rows for each head, and $^*$ marks a significant paired difference. ES is the energy score, and Penetr.\ is the share of samples that penetrate the table.}
\end{table}

\subsection{Latent Regularization Loss}
\label{sec:app_reg}

Points2Plans~\citep{24-arxiv-points2plans} trains its deterministic decoder with the latent transition loss $\mathcal{L}_{\mathrm{reg}}$ (Section~\ref{sec:training}). We trained each probabilistic head with the same loss and evaluated it on the horizontal-push test set (Table~\ref{tab:reg}), with $\lambda_r$ chosen per head on the validation set. For every head except state tokenization, the loss significantly worsened the energy scores, the Brier score, and plausibility, and for the MDN and the Gaussian it raised the fraction of samples that penetrate the table to about 48\%. The much lower calibration error of the MDN and the Gaussian reflects predictions that are spread widely enough to cover the outcome, not more accurate ones. For state tokenization, the loss lowered the position energy score but left the Brier score and plausibility unchanged.

We attribute this to a mismatch between the loss and the probabilistic heads. The loss is identical for every head. It pulls the predicted next latent $\mathbf{z}_t+\delta\mathbf{z}_t$ toward the latent of the observed next state, $\Enc_s(S_{t+1})$. This suits the deterministic model, which predicts a single outcome. A probabilistic head, however, needs a latent that represents every outcome the action can produce, and the loss penalizes this latent for not matching the single observed one. We therefore train the probabilistic heads without $\mathcal{L}_{\mathrm{reg}}$, including state tokenization for consistency, and keep it for the deterministic baseline to follow Points2Plans. A latent transition loss suited to stochastic outcomes, such as matching the predicted latent to a distribution of next encodings, is left for future work.

\begin{figure*}[tp]
\centering
\includegraphics[width=\textwidth]{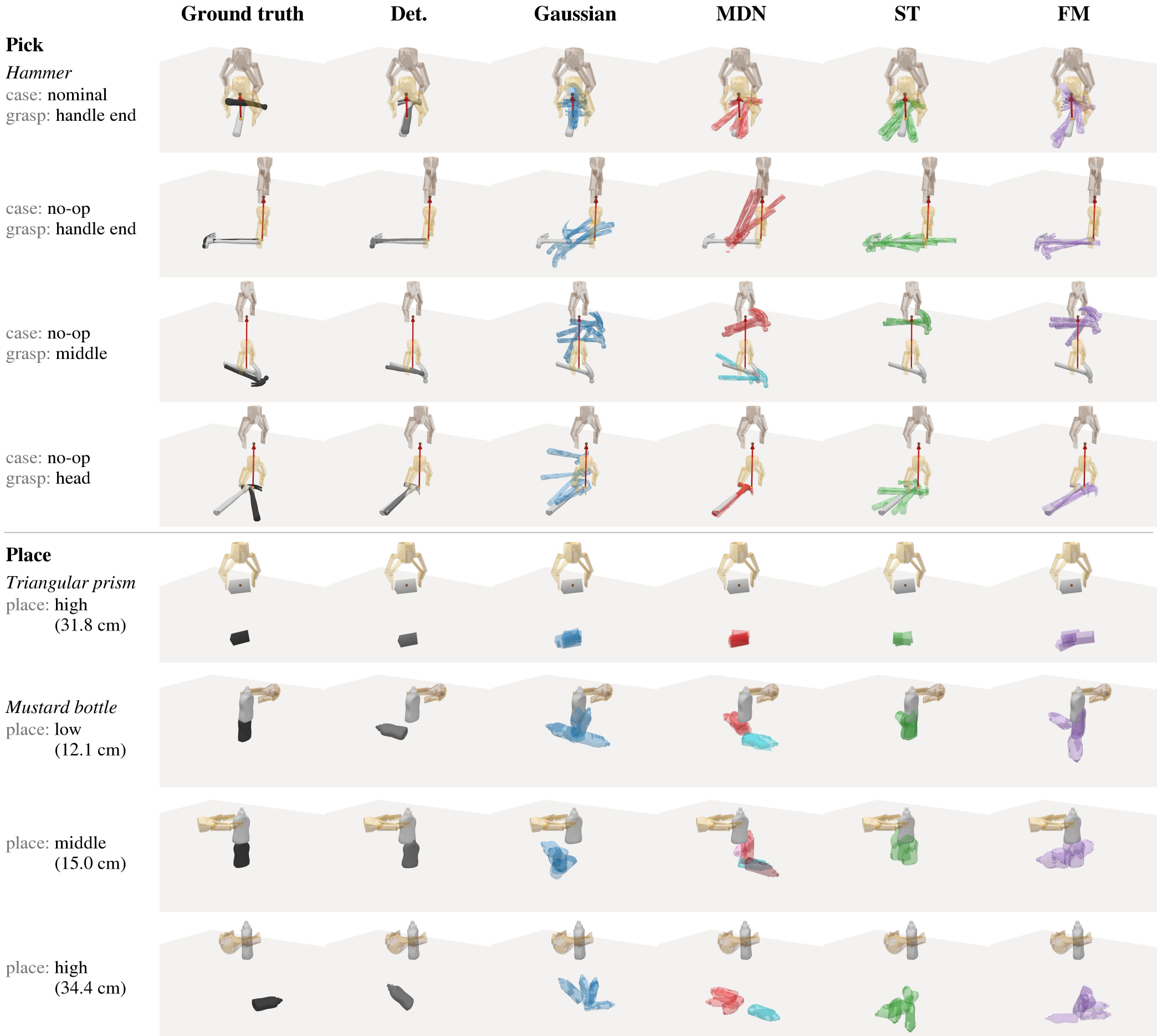}
\caption{Additional pick and place predictions. Each row is one test episode (five samples per stochastic model; MDN colored by mixture component).}
\label{fig:qual_app_pp}
\end{figure*}

\subsection{Planner Details}
\label{sec:app_planner}

Algorithm~\ref{alg:cem} details the planner of Section~\ref{sec:planning}. Each candidate is rolled out with $M$ outcome samples. At each step, its skill is fixed once in the world frame and then expressed in the object frame of each sample, so all samples of a candidate execute the same skill. The sampled relative pose is then applied to the object pose and point cloud of that sample before the next skill is evaluated.

\begin{algorithm}[t]
\footnotesize
\caption{Monte Carlo (MC)-style CEM planner}\label{alg:cem}
\begin{algorithmic}[1]
\Require $N_{\mathrm{cand}}$ candidates, $N_{\mathrm{elite}}$ elites, $M$ outcome samples, $N_{\mathrm{iter}}$ iterations, horizon $H$, skills $a_{0:H-1}$, initial observation $S_0=\{O_0,E_0\}$, initial object pose $p_0$, goal distribution $P_G$
\State \textbf{global:} $\Enc_s$, $\Enc_a$, $\Phi$, $\Dec_\psi$, $\omega$, $\mathrm{Ref}$, $\mathrm{World}$, $\mathrm{Local}$
\Function{\texttt{MC\_CEM}}{$S_0, p_0, P_G$}
    \State Initialize $\pi_1$ to the action prior
    \For{$k=1,\dots,N_{\mathrm{iter}}$}
        \State $\boldsymbol{\theta}^{(i)}_{0:H-1} \sim \pi_k$ for $i=1,\dots,N_{\mathrm{cand}}$
        \State $\tau^{(i)}=(A^{(i)}_0,\ldots,A^{(i)}_{H-1})$ with $A^{(i)}_t=(a_t,\boldsymbol{\theta}^{(i)}_t)$
        \State Feasibility check on each $\tau^{(i)}$
        \State $O^{(i,m)}_0=O_0$, \quad $p^{(i,m)}_0=p_0$ for $m=1,\dots,M$
        \For{$t=0,\dots,H-1$}
            \State $W_t^{(i)}=\mathrm{World}\big(A_t^{(i)},\mathrm{Ref}(\{O^{(i,m)}_t\}_{m})\big)$
            \State $A^{(i,m)}_t=\mathrm{Local}\big(W_t^{(i)},O^{(i,m)}_t\big)$
            \State $S^{(i,m)}_t=\{O^{(i,m)}_t,E_0\}$
            \State $\delta \mathbf{z}_t^{(i,m)}=\Phi\big(\Enc_s(S_t^{(i,m)}),\Enc_a(A_t^{(i,m)})\big)$
            \State $\delta p_t^{(i,m)} \sim \Dec_\psi(\delta \mathbf{z}_t^{(i,m)})$
            \State $p_{t+1}^{(i,m)}=p_t^{(i,m)}\,\delta p_t^{(i,m)}$
            \State $O_{t+1}^{(i,m)}=\omega(\delta p_t^{(i,m)})\,O_t^{(i,m)}$
        \EndFor
        \State $\widehat J(\tau^{(i)})=\frac{1}{M}\sum_{m=1}^{M}c\big(p^{(i,m)}_{H},P_G\big)$ for $i=1,\dots,N_{\mathrm{cand}}$
        \State $\pi_{k+1} \gets$ refit to the $N_{\mathrm{elite}}$ lowest-cost $\boldsymbol{\theta}^{(i)}_{0:H-1}$
    \EndFor
    \State \Return the lowest-cost $\tau^{(i)}$
\EndFunction
\end{algorithmic}
\end{algorithm}

\subsection{Additional Prediction Results}
\label{sec:app_qual}

\begin{table}[t]
\centering
\caption{Paired difference in the outcome-class Brier score.}
\label{tab:brier_diff}
\footnotesize
\setlength{\tabcolsep}{2pt}
\newcommand{\cell}[2]{\makecell{#1\\[-1pt]{\scriptsize[#2]}}}
\resizebox{\columnwidth}{!}{%
\begin{tabular}{lcccc}
\toprule
Model A \(-\) model B & \makecell{Horizontal\\push} & \makecell{Vertical\\push} & Pick & Place \\
\midrule
MDN \(-\) Det. & \cell{\textbf{\(-\)0.187}}{\(-\)0.263, \(-\)0.113} & \cell{\textbf{\(-\)0.251}}{\(-\)0.316, \(-\)0.182} & \cell{\textbf{\(-\)0.281}}{\(-\)0.326, \(-\)0.232} & \cell{\textbf{\(-\)0.289}}{\(-\)0.380, \(-\)0.191} \\
\addlinespace[2pt]
ST \(-\) Det. & \cell{\textbf{\(-\)0.214}}{\(-\)0.304, \(-\)0.139} & \cell{\textbf{\(-\)0.299}}{\(-\)0.370, \(-\)0.226} & \cell{\textbf{\(-\)0.185}}{\(-\)0.224, \(-\)0.145} & \cell{\textbf{\(-\)0.359}}{\(-\)0.450, \(-\)0.264} \\
\addlinespace[2pt]
FM \(-\) Det. & \cell{\textbf{\(-\)0.281}}{\(-\)0.373, \(-\)0.199} & \cell{\textbf{\(-\)0.290}}{\(-\)0.361, \(-\)0.213} & \cell{\textbf{\(-\)0.353}}{\(-\)0.410, \(-\)0.293} & \cell{\textbf{\(-\)0.460}}{\(-\)0.551, \(-\)0.372} \\
\midrule
MDN \(-\) Gaussian & \cell{\textbf{\(-\)0.180}}{\(-\)0.225, \(-\)0.136} & \cell{\textbf{\(-\)0.334}}{\(-\)0.453, \(-\)0.220} & \cell{\textbf{\(-\)0.529}}{\(-\)0.577, \(-\)0.487} & \cell{\textbf{\(-\)0.299}}{\(-\)0.404, \(-\)0.205} \\
\addlinespace[2pt]
ST \(-\) Gaussian & \cell{\textbf{\(-\)0.207}}{\(-\)0.294, \(-\)0.128} & \cell{\textbf{\(-\)0.382}}{\(-\)0.494, \(-\)0.272} & \cell{\textbf{\(-\)0.434}}{\(-\)0.492, \(-\)0.378} & \cell{\textbf{\(-\)0.369}}{\(-\)0.477, \(-\)0.272} \\
\addlinespace[2pt]
FM \(-\) Gaussian & \cell{\textbf{\(-\)0.274}}{\(-\)0.360, \(-\)0.193} & \cell{\textbf{\(-\)0.374}}{\(-\)0.484, \(-\)0.260} & \cell{\textbf{\(-\)0.601}}{\(-\)0.643, \(-\)0.563} & \cell{\textbf{\(-\)0.471}}{\(-\)0.579, \(-\)0.375} \\
\midrule
FM \(-\) ST & \cell{\textbf{\(-\)0.067}}{\(-\)0.097, \(-\)0.040} & \cell{+0.009}{\(-\)0.018, 0.038} & \cell{\textbf{\(-\)0.167}}{\(-\)0.203, \(-\)0.135} & \cell{\textbf{\(-\)0.102}}{\(-\)0.128, \(-\)0.074} \\
\addlinespace[2pt]
MDN \(-\) ST & \cell{+0.027}{\(-\)0.035, 0.092} & \cell{\textbf{+0.048}}{0.014, 0.082} & \cell{\textbf{\(-\)0.096}}{\(-\)0.125, \(-\)0.068} & \cell{\textbf{+0.070}}{0.040, 0.104} \\
\midrule
Gaussian \(-\) Det. & \cell{\(-\)0.006}{\(-\)0.087, 0.081} & \cell{+0.084}{\(-\)0.019, 0.189} & \cell{\textbf{+0.248}}{0.177, 0.322} & \cell{+0.010}{\(-\)0.052, 0.070} \\
\bottomrule
\end{tabular}%
}\\[2pt]
\parbox{\columnwidth}{\footnotesize Model A minus model B, so negative values favor model A. Brackets show 95\% object-level bootstrap intervals, and bold marks a significant difference.}
\end{table}

\begin{figure*}[tp]
\centering
\includegraphics[width=\textwidth]{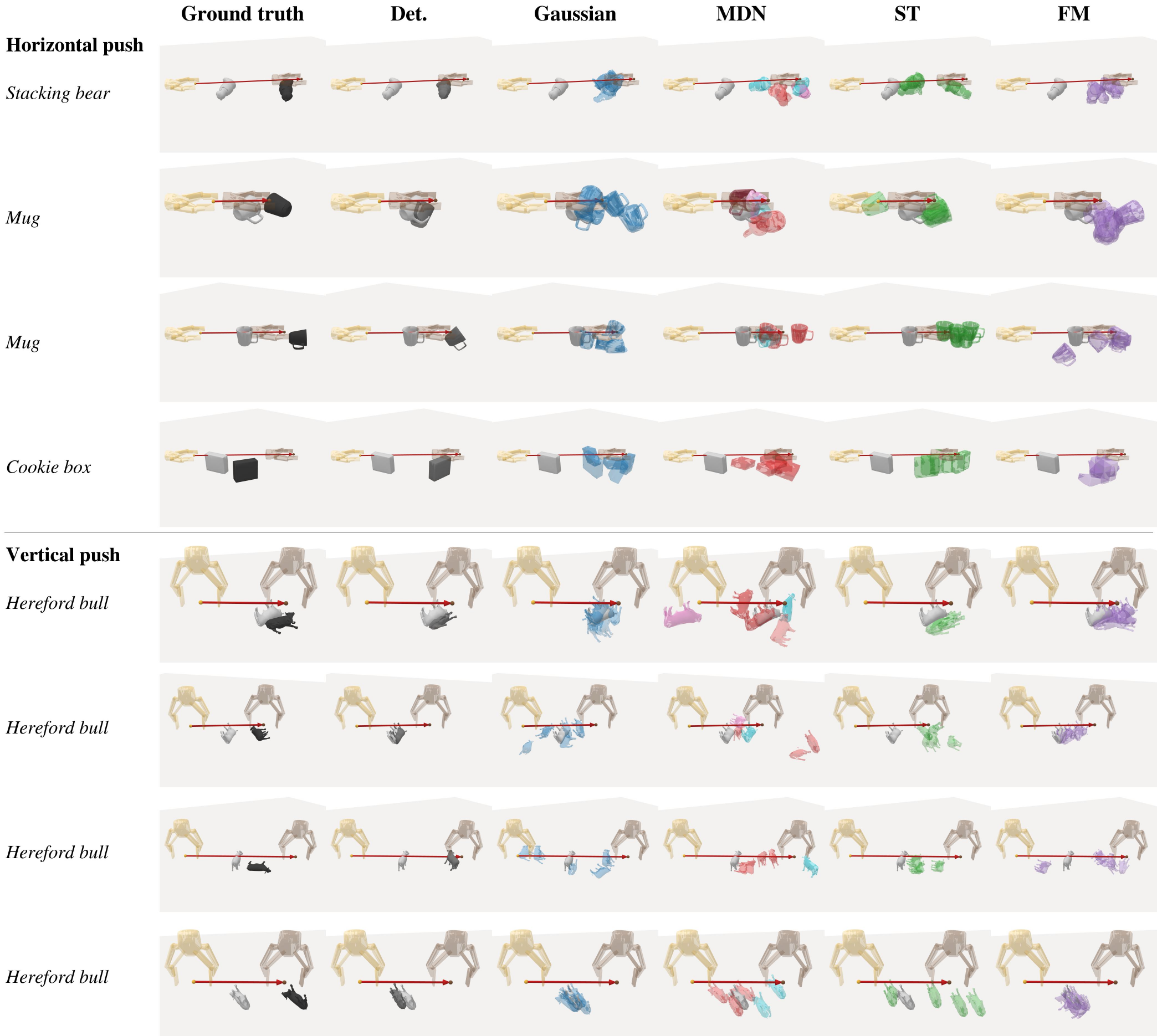}
\caption{Additional push predictions. Each row is one test episode (five samples per stochastic model; MDN colored by mixture component).}
\label{fig:qual_app_push}
\end{figure*}

Figures~\ref{fig:qual_app_push} and~\ref{fig:qual_app_pp} show additional test episodes, in which each row varies one condition, such as the start pose, the grasp location, or the release height. Near the boundary between outcomes, the Gaussian spreads both the position and the orientation of the object. On push and pick, its samples often trail along the direction of the push or the lift, or take orientations that are not rest poses. The deterministic model instead averages across the boundary, so its prediction is often not a rest pose or lies at the wrong position, and it cannot represent the other outcome when the outcomes split. Both models show the same failure on place, particularly at higher release heights, where the bottle can land either upright or toppled. The multimodal models occasionally struggle, but they predict rest poses in most episodes and capture the split between outcomes. On hard no-op picks in particular, they rarely predict a lift when the object stays on the table.

\begin{table}[t]
\centering
\caption{Outcome-class Brier score on seen and unseen test objects.}
\label{tab:brier_seen_unseen}
\footnotesize
\setlength{\tabcolsep}{3pt}
\begin{tabular}{llccccc}
\toprule
Skill & Split & Det. & Gaussian & MDN & ST & FM \\
\midrule
\multirow{2}{*}{\makecell[l]{Horizontal\\push}} & Seen & 0.580 & 0.538 & 0.371 & 0.351 & \textbf{0.290} \\
 & Unseen & 0.793 & 0.826 & 0.631 & 0.596 & \textbf{0.523} \\
\midrule
\multirow{2}{*}{\makecell[l]{Vertical\\push}} & Seen & 0.620 & 0.748 & 0.351 & 0.286 & \textbf{0.273} \\
 & Unseen & 0.621 & 0.666 & 0.387 & \textbf{0.354} & 0.381 \\
\midrule
\multirow{2}{*}{Pick} & Seen & 0.718 & 0.999 & 0.455 & 0.554 & \textbf{0.384} \\
 & Unseen & 0.902 & 1.113 & 0.601 & 0.692 & \textbf{0.527} \\
\midrule
\multirow{2}{*}{Place} & Seen & 0.957 & 0.999 & 0.721 & 0.665 & \textbf{0.574} \\
 & Unseen & 1.132 & 1.106 & 0.782 & 0.697 & \textbf{0.583} \\
\bottomrule
\end{tabular}\\[2pt]
\parbox{\columnwidth}{\footnotesize Lower is better. Bold marks the lowest value in each row.}
\end{table}

Tables~\ref{tab:brier_diff} and~\ref{tab:brier_seen_unseen} report the outcome-class Brier score in detail, and Table~\ref{tab:seed_eval} repeats the prediction metrics on every skill across three training seeds. In every seed and on every skill, all three multimodal decoders achieved a lower Brier score than both the deterministic model and the Gaussian. The Gaussian had a higher Brier score than the deterministic model in all three seeds on pick, place, and vertical push, and a lower one in all three on horizontal push.

\subsection{Additional Task Success Results}
\label{sec:app_success}

\textbf{Replay comparison:} Replaying the demonstrated action shows how much of each task is achievable. It succeeded in 79.7\% of the horizontal-push episodes and 83.7\% of the vertical-push episodes at $(5\,\mathrm{cm},20^\circ)$, whereas the best planned actions reached 43.8\% and 44.2\%. For place, replaying the demonstrated grasp lifted the object in 92\% of the episodes. Place success was measured on these episodes only, so it reflects the release and not the preceding pick. Because the replayed grasp occasionally failed in different episodes across runs, we also restricted the comparison to the episodes in which it succeeded for every model. The differences between models changed by less than one percentage point. Even on push, the replayed action ended in a rest pose different from the recorded one in 4.9--8.2\% of the episodes, which shows that the skill outcomes are stochastic.

\begin{figure*}[tp]
\centering
\includegraphics[width=\textwidth]{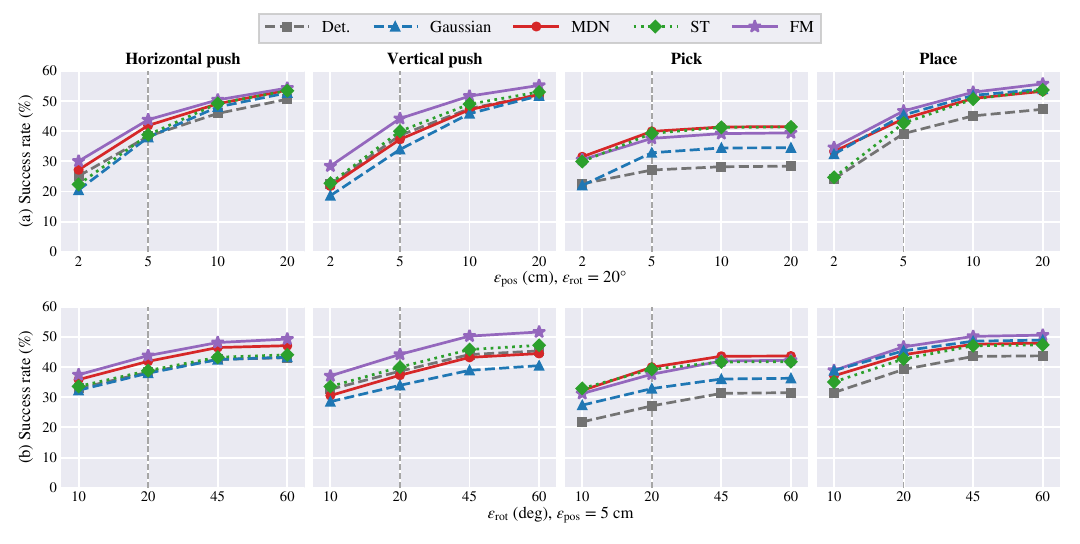}
\caption{Simulated task success over the success tolerances of Eq.~\eqref{eq:success}. (a) Success versus the position tolerance $\epsilon_{\mathrm{pos}}$ at $\epsilon_{\mathrm{rot}}=20^\circ$. (b) Success versus the rotation tolerance $\epsilon_{\mathrm{rot}}$ at $\epsilon_{\mathrm{pos}}=5$\,cm. Dashed vertical lines mark the main tolerance used in the text. ST and FM denote state tokenization and flow matching.}
\label{fig:task_success_tol}
\end{figure*}

\textbf{Tolerance grid:} The main tolerance of 5\,cm is about three times the median error of the object pose estimated from a partial point cloud (1.5\,cm), and $20^\circ$ is twice the yaw standard deviation of the goal distribution. Figure~\ref{fig:task_success_tol} varies each tolerance around the main setting, and the ordering of the models largely holds across tolerances.

\textbf{Seen and unseen objects:} On unseen objects (Figure~\ref{fig:success_seen_unseen}), flow matching remained significantly better than the deterministic model on every skill, and the MDN did so on horizontal push, pick, and place.

\textbf{Real robot:} We segmented the object from a single RGB-D view by fitting the table plane with RANSAC~\citep{ransac}. For objects with a mesh, we estimated the pose with ICP~\citep{icp}, initialized from each rest pose of the object, and used an AprilTag only as a fallback when ICP was ambiguous or the object was occluded. For objects without a mesh, we used the point cloud's footprint for position and heading and its height for the rest pose. We checked every result against the camera video and corrected the estimate where it disagreed, applying the same rule to all models. Because an object's rest poses on the real table can differ slightly from those in simulation, each goal used the tilt of the corresponding real rest pose.

Push and place used the same criteria as in simulation at $(5\,\mathrm{cm},20^\circ)$. Pick counted a task as successful if the object was held and lifted, without the pose tolerance. Under the full criterion of Eq.~\eqref{eq:success}, pick success was 20.0\% (deterministic), 46.7\% (Gaussian), 53.3\% (MDN), 26.7\% (state tokenization), and 53.3\% (flow matching).

\subsection{Inference and Planning Timing Evaluations}
\label{sec:app_timing}

\begin{table}[t]
\centering
\caption{Inference and planning time.}
\label{tab:timing}
\footnotesize
\setlength{\tabcolsep}{3pt}
\begin{tabular}{lcccccc}
\toprule
 & \multicolumn{2}{c}{Inference (ms)} & \multicolumn{4}{c}{Planning (s)} \\
\cmidrule(lr){2-3}\cmidrule(lr){4-7}
Model & Single & Batch$^\dagger$ & $N_{\mathrm{cand}}{=}100$ & 250 & 500 & 1000 \\
\midrule
Det. & 3.4 & 5.6 & 0.32 & 0.33 & 0.36 & 0.41 \\
Gaussian & 6.0 & 37.5 & 0.49 & 0.81 & 1.24 & 1.50 \\
MDN & 9.3 & 58.5 & 0.59 & 0.84 & 1.35 & 2.15 \\
ST & 4.8 & 975 & 2.49 & 5.83 & 11.2 & 23.6 \\
FM & 46.5 & 1711 & 5.38 & 12.7 & 24.9 & 48.9 \\
\bottomrule
\end{tabular}\\[2pt]
\parbox{\columnwidth}{\footnotesize Median times. Single is one sample for one start state. $^\dagger$1000 candidates $\times$ 100 samples (1 sample for Det.). Planning uses 10 CEM iterations with $N_{\mathrm{cand}}$ candidates.}
\end{table}

We measured sample generation and planning time on one NVIDIA H200 GPU and report the median (Table~\ref{tab:timing}). For inference, each model predicted outcomes for one start state, and we also evaluated a CEM-sized batch of 1000 candidate actions with 100 samples each. For planning, we ran CEM with 10 iterations and $N_{\mathrm{cand}}\in\{100,250,500,1000\}$ candidates for one arm and one goal over 21 horizontal-push starts, excluding inverse kinematics. The deterministic model was the fastest, and both the MDN and the Gaussian planned within about 2\,s even with 1000 candidates. Flow matching was the slowest per sample because each sample is integrated over five midpoint steps. With many candidates, state tokenization and flow matching took an order of magnitude longer to plan than the MDN and the Gaussian.

\end{document}